\documentclass[nohyperref]{article}

\usepackage{microtype}
\usepackage{graphicx}
\usepackage{booktabs} 

\usepackage{hyperref}

\usepackage[accepted]{icml2022}

\usepackage{amsmath}
\usepackage{amssymb}
\usepackage{mathtools}
\usepackage{amsthm}

\usepackage[capitalize,noabbrev]{cleveref}

\theoremstyle{plain}

\theoremstyle{definition}

\theoremstyle{remark}

\usepackage[textsize=tiny]{todonotes}

\icmltitlerunning{Quantifying and Learning Linear Symmetry-Based Disentanglement}

\usepackage[justification=centering]{subcaption}
\usepackage{multirow}

\newcommand{\methodname}{LSBD-VAE}
\newcommand{\metricname}{$\mathcal{D}_\mathrm{LSBD}$}

\begin{document}

\twocolumn[
\icmltitle{Quantifying and Learning Linear Symmetry-Based Disentanglement}



\icmlsetsymbol{equal}{*}

\begin{icmlauthorlist}
\icmlauthor{Loek Tonnaer}{equal,tue,eaisi}
\icmlauthor{Luis A. P\'erez Rey}{equal,tue,eaisi,prosus}
\icmlauthor{Vlado Menkovski}{tue,eaisi}
\icmlauthor{Mike Holenderski}{tue,eaisi}
\icmlauthor{Jacobus W. Portegies}{tue,eaisi}
\end{icmlauthorlist}

\icmlaffiliation{tue}{Eindhoven University of Technology (TU/e), Eindhoven, The Netherlands}
\icmlaffiliation{eaisi}{Eindhoven Artificial Intelligence Systems Institute (EAISI), Eindhoven, the Netherlands}
\icmlaffiliation{prosus}{Prosus, Amsterdam, The Netherlands}

\icmlcorrespondingauthor{Loek Tonnaer}{l.m.a.tonnaer@tue.nl}
\icmlcorrespondingauthor{Luis A. P\'erez Rey}{l.a.perez.rey@tue.nl}

\icmlkeywords{Machine Learning, ICML, Disentanglement}

\vskip 0.3in
]



\printAffiliationsAndNotice{\icmlEqualContribution} 

\begin{abstract}
The definition of Linear Symmetry-Based Disentanglement (LSBD) formalizes the notion of linearly disentangled representations, but there is currently no metric to quantify LSBD. Such a metric is crucial to evaluate LSBD methods and to compare to previous understandings of disentanglement. We propose \metricname{}, a mathematically sound metric to quantify LSBD, and provide a practical implementation for $\mathrm{SO}(2)$ groups. Furthermore, from this metric we derive \methodname{}, a semi-supervised method to learn LSBD representations. We demonstrate\footnote{Code available at \url{https://github.com/luis-armando-perez-rey/lsbd-vae}} the utility of our metric by showing that (1) common VAE-based disentanglement methods don't learn LSBD representations, (2) \methodname{}, as well as other recent methods, \emph{can} learn LSBD representations needing only limited supervision on transformations, and (3) various desirable properties expressed by existing disentanglement metrics are also achieved by LSBD representations.
\end{abstract}

\section{Introduction}
Learning low-dimensional representations that disentangle the underlying factors of variation in data is considered an important step towards interpretable machine learning with good generalization. To address the fact that there is no consensus on what disentanglement entails and how to formalize it, \citet{higgins2018towards} propose a formal definition for Linear Symmetry-Based Disentanglement, or LSBD, arguing that underlying real-world symmetries give exploitable structure to data (see Sect.~\ref{sec:lsbd}).

LSBD emphasizes that the variability in data observations is often due to some transformations, and that good data representations should reflect these transformations. A typical setting is that of an agent interacting with its environment. An action of the agent will transform some aspect of the environment and its observation thereof, but keeps all other aspects invariant. It is often easy and cheap to register the actions that an agent performs and how they transform the observed environment, which can provide useful information for learning disentangled representations.

However, there is currently no general metric to quantify LSBD. Such a metric is crucial to properly evaluate methods aiming to learn LSBD representations and to relate LSBD to previous definitions of disentanglement. Although previous works have evaluated LSBD by measuring performance on downstream tasks~\citep{Caselles-Dupre2019} or by measuring specific traits related to LSBD~\citep{painter2020linear, Quessard2020}, none of these evaluation methods directly quantify LSBD according to its formal definition.

We propose \metricname{}, a well-formalized and generally applicable metric that quantifies the level of LSBD in learned data representations (Sect.~\ref{sec:d_lsbd}). We show an intuitive justification of this metric, as well as its theoretical derivation. We also provide a practical implementation to compute \metricname{} for common $\mathrm{SO}(2)$ symmetry groups. Furthermore, we show that our metric formulation can be used to derive a semi-supervised method to learn LSBD representations, which we call \methodname{} (Sect.~\ref{sec:lsbd-vae}). To make \methodname{} more widely applicable, we also demonstrate how to disentangle symmetric properties from other non-symmetric properties, and how to quantify this disentanglement with \metricname{}.

We show the utility of \metricname{} by quantifying LSBD in a number of settings, for a variety of datasets with underlying $\mathrm{SO}(2)$ symmetries and other non-symmetric properties (Sect.~\ref{sec:experimental_setup} \& \ref{sec:results}). First, we evaluate common VAE-based disentanglement methods and show that most don't learn LSBD representations. Second, we evaluate \methodname{} and other recent methods that specifically target LSBD, showing that they \emph{can} obtain much better \metricname{} scores while needing only limited supervision on transformations. Third, we compare \metricname{} with existing disentanglement metrics, showing that various desirable properties expressed with these metrics are also achieved by LSBD representations.

\section{Related Work}
Plenty of works have focused on learning and quantifying disentangled representations recently, but research has shown that there is little consensus about the exact definition of disentanglement and methods often do not achieve it as well as they proclaim~\citep{locatello2019challenging}. To introduce some much-needed formalization, \citet{higgins2018towards} proposed to define disentanglement with respect to symmetry transformations acting on the data. They used group theory to provide two formal definitions, which we refer to as (Linear) Symmetry-Based Disentanglement, or (L)SBD. In this paper we focus only on LSBD, not SBD.

Several methods have been proposed to learn LSBD representations~\citep{Caselles-Dupre2019, painter2020linear, Quessard2020}. These methods also learn to represent the transformations acting on the input data, assuming various levels of supervision on these transformations. Other methods have previously focused on capturing transformations of the data outside the context of disentanglement as well~\citep{Cohen2015, Sosnovik2019, Worrall2017a}.

Although some of these works do propose metrics that measure some aspect of LSBD, none of them provide a general metric that directly quantifies LSBD according to its formal definition and for any data representation. \citet{painter2020linear} mention two metrics:  \emph{Independence Score} measures whether the actions of the subgroups have effects on independent vector spaces, \emph{Factor Leakage} only measures the number of dimensions in which the subgroup actions are encoded, which is not a property required by LSBD. Neither are general quantifications of LSBD. Additionally, \citet{Quessard2020} also propose a ``metric'', but this is in fact a loss component particular to their group representation parameterization and cannot be used as a general metric for LSBD.

\section{Linear Symmetry-Based Disentanglement}
\label{sec:lsbd}

\citet{higgins2018towards} provide a formal definition of linear disentanglement that connects symmetry transformations affecting the real world (from which data is observed) to the internal representations of a model. The definition is grounded in concepts from \emph{group theory}, we provide a more detailed description of these concepts in Appendix~\ref{app:preliminaries}.

The definition\footnote{The original definition actually considers an additional set of world states $W$, but our definition is more practical and can be shown to be the same under mild conditions, see Appendix~\ref{app:world_states}.} considers a group $G$ of symmetry transformations acting on the \emph{data space} $X$ through the group action $\cdot: G\times X\rightarrow X$. In particular, $G$ can be decomposed as the direct product of $K$ groups $G = G_1\times\ldots\times G_K$. A model's internal representation of data is modeled with the \emph{encoding} function $h:X\rightarrow Z$ that maps data to the \emph{embedding space} $Z$. The definition for Linearly Symmetry-Based Disentangled (LSBD) representations then formalizes the requirement that a model's encoding $h$ should reflect and disentangle the transformation properties of the data, and that the transformation properties of the model's encoding should be linear. The exact definition is as follows:

\paragraph{Definition: Linear Symmetry-Based Disentanglement (LSBD)}
A model's encoding map $h:X\rightarrow Z$, where $Z$ is a vector space, is LSBD with respect to the group decomposition $G = G_1\times\ldots\times G_K$ if
\begin{enumerate}
    \item there is a decomposition of the embedding space $Z = Z_1\oplus \ldots\oplus Z_K$ into $K$ vector subspaces,
    \item there are group representations for each subgroup in the corresponding vector subspace $\rho_k: G_k\rightarrow \mathrm{GL}(Z_k)$, $k\in\{1,\ldots,K\}$
    \item the group representation $\rho:G \rightarrow \mathrm{GL}(Z)$ acts on $Z$ as
    \begin{equation}
        \rho(g)\cdot z = (\rho_1(g_1)\cdot z_1,\ldots, \rho_K(g_K)\cdot z_K),
    \end{equation}
     for $g = (g_1, \ldots, g_K)\in G$ and $z= (z_1,\ldots, z_K)\in Z$ with $g_k\in G_k$ and $z_k\in Z_k$.
    \item the map $h$ is \emph{equivariant} with respect to the actions of $G$ on $X$ and $Z$, i.e. , for all $x\in X$ and $g\in G$ it holds that $h(g\cdot x) = \rho(g)\cdot h(x)$.
\end{enumerate}

Furthermore, we say that a group representation $\rho$ is \emph{linearly disentangled} with respect to the group decomposition $G = G_1\times\ldots\times G_K$ if it satisfies criteria 1 to 3 from the LSBD definition above.

\begin{figure}[ht!]
    \includegraphics[width=1.0\linewidth]{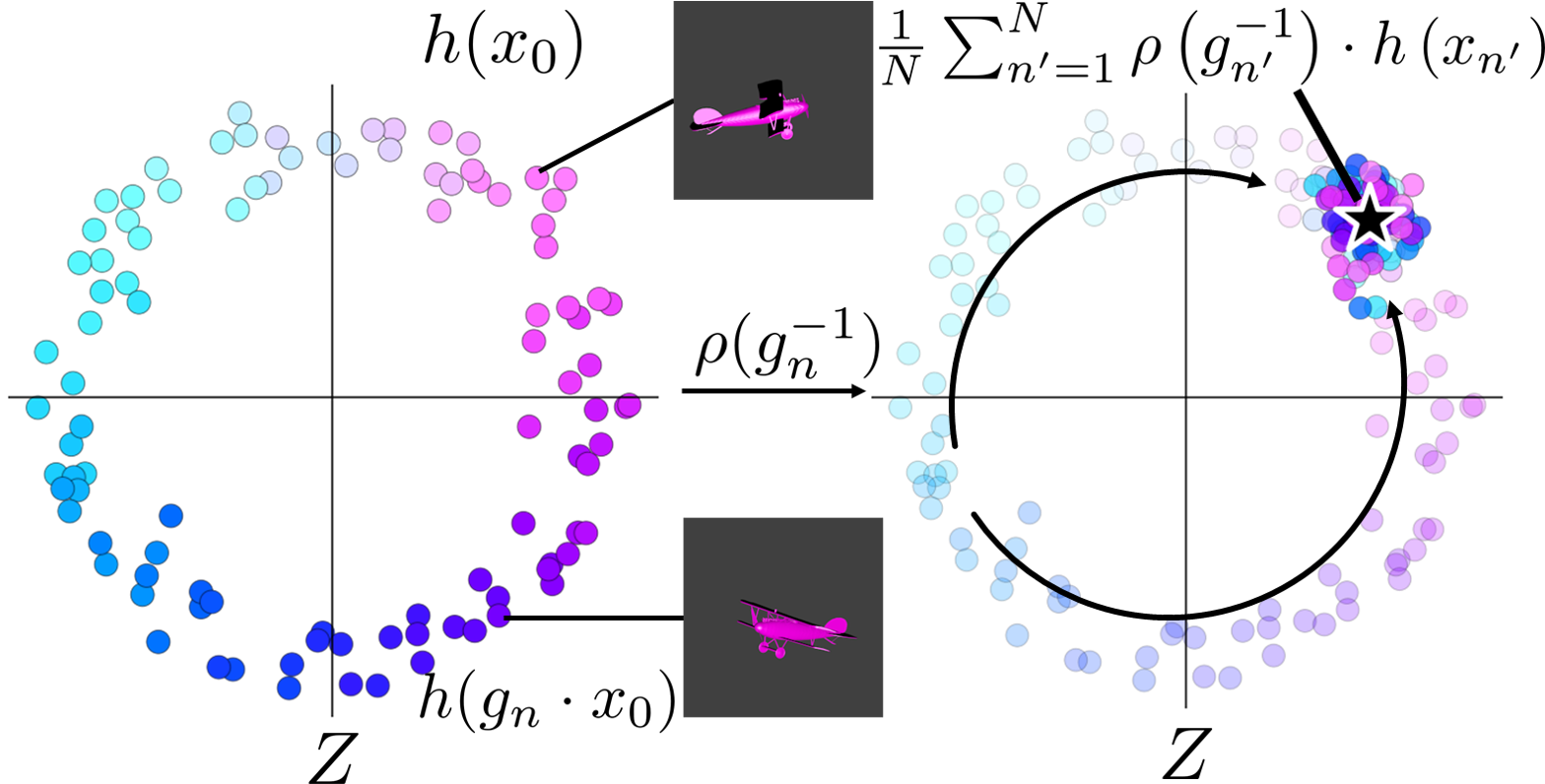}
    \caption{A dataset of images from a rotating object expressed in terms of the group $G = \mathrm{SO}(2)$ acting on a base image $x_0$. It is possible to quantify the level of LSBD of an encoding map $h$ by measuring its equivariance with respect to a group representation $\rho$. Since all data has been generated from $x_0$, equivariance can be measured as the dispersion of the points $\{ \rho(g_n^{-1})\cdot h(x_n) \}_{n=1}^N$.}
    \label{fig:diagram_quantifying1}
\end{figure}

\begin{figure*}[t!]

\centering
  \includegraphics[width=0.9\linewidth]{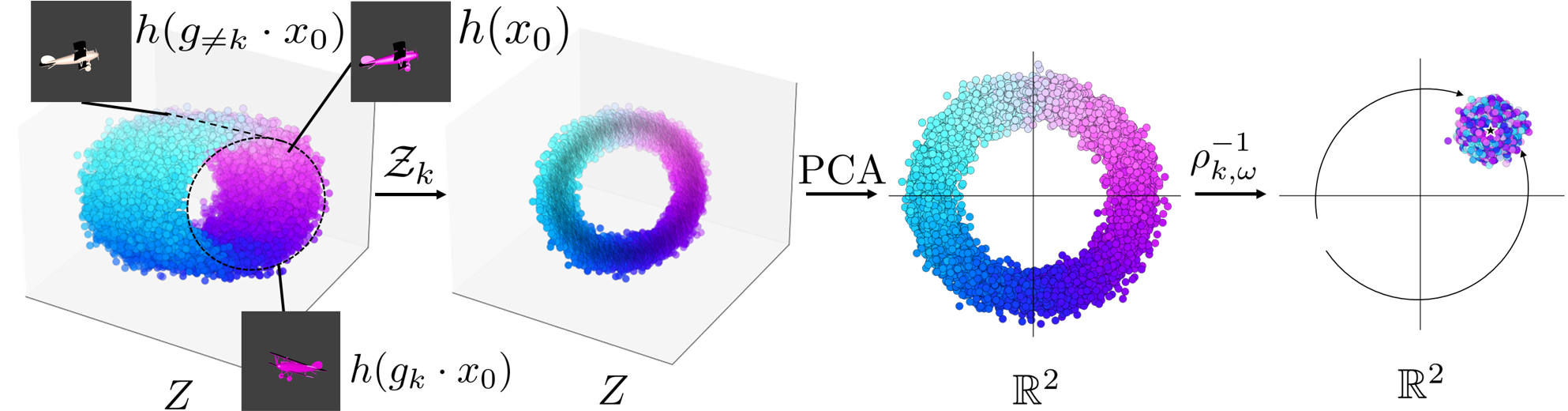}

  \caption{Consider a dataset modeled by a group decomposition $G = G_1\times\cdots\times G_K$ acting on $x_0$ and embedded in a latent space $Z$ via $h$. In this example the subgroup $G_k = \mathrm{SO}(2)$ models the rotations of an airplane. Other subgroups $G_{\neq k}$ could also be acting e.g. changes in airplane color. The first step to calculate the disentanglement of $G_k$ is to construct a set of data embeddings $\mathcal{Z}_k\subseteq Z$ whose variability is due to $G_k$. These embeddings are then projected into a 2-dimensional space through PCA. For these projected embeddings we can describe the group representations in a simple parametric form $\rho_{k,w}$. For a given $\rho_{k,w}$ the equivariance of $G_k$  is measured as the dispersion after applying the action of the inverse group representation $\rho_{k,w}^{-1}$. }
  \label{fig:diagram_quantifying2}
\end{figure*}

\section{Quantifying LSBD: \metricname{}}
\label{sec:d_lsbd}
\subsection{Intuition: Measuring Equivariance with Dispersion}

To motivate our metric, let's first assume a setting in which a suitable \emph{linearly disentangled} group representation $\rho$ is known. Let's further assume that the dataset of observations can be expressed with respect to $G$ acting on some base point $x_0 \in X$, i.e. $\{x_n\}_{n=1}^N = \{g_n \cdot x_0\}_{n=1}^N$. Formally, this assumes that the action of $G$ on $X$ is \emph{regular}. In this case, we can use the inverse group elements $g_n^{-1}$ to transform each data point toward the base point $x_0$, i.e.
\begin{align}
    \label{eq:transform_to_basepoint}
    x_0 = g_1^{-1}\cdot x_1 = \ldots = g_N^{-1}\cdot x_N.
\end{align}

Since $\rho$ is \emph{linearly disentangled}, we only need to measure the \emph{equivariance} of the encoding map $h$ to quantify LSBD. Equivariance is achieved when $h(g\cdot x) = \rho(g)\cdot h(x)$, for all $g\in G, x\in X$. Given the dataset described above, we can check this property for $x \in \{x_n\}_{n=1}^N$ and $g \in \{g_n\}_{n=1}^N$.\footnote{Note that $\{g_n\}_{n=1}^N$ can be used to describe all known group transformations between elements in the dataset by means of composition and inverses, since $x_i=g_i\cdot(g_j^{-1}\cdot x_j)$. Thus it suffices to check equivariance for these $N$ group transformations.} In particular, from Equation~\eqref{eq:transform_to_basepoint} we can see that we have equivariance if
\begin{align}
    h(x_0) = \rho(g_1^{-1})\cdot h(x_1) = \ldots = \rho(g_N^{-1})\cdot h(x_N).
\end{align}

This not only characterizes perfect equivariance, but also allows for an efficient way to quantify how close we are to true equivariance, by measuring the \emph{dispersion} of the points $\{ \rho(g_n^{-1})\cdot h(x_n) \}_{n=1}^N$.\footnote{Note that we do not actually need to know $x_0$ nor $h(x_0)$.} Given a suitable norm $\|\cdot\|_Z$ in $Z$, we can thus quantify LSBD in this setting as
\begin{equation}
\begin{aligned}
\label{eq:simple_metric}
    \frac{1}{N}\sum_{n=1}^N \left\| \rho(g_n^{-1})\cdot h(x_n) - M^* \right\|_Z^2, \\
    \mbox{with} \; M^* = \frac{1}{N}\sum_{n'=1}^N \rho(g_{n'}^{-1})\cdot h(x_{n'}),
\end{aligned}
\end{equation}
i.e. we compute the mean $M^*$ of $\{ \rho(g_n^{-1})\cdot h(x_n) \}_{n=1}^N$ and use the average squared distance to this mean for points in $\{ \rho(g_n^{-1})\cdot h(x_n) \}_{n=1}^N$ as our LSBD metric, see Fig.~\ref{fig:diagram_quantifying1}.

However, this formulation requires knowing the right \emph{linearly disentangled} group representation and a suitable norm in $Z$. Moreover, it implicitly assumes a uniform probability measure over the group elements $\{g_n\}_{n=1}^N$. In the next section we formulate our metric for a more general setting.

\subsection{\metricname{}: A Metric for LSBD}

Generalizing the ideas from the previous section with concepts from \emph{measure theory}, we propose a metric to measure the level of LSBD of any encoding $h:X \rightarrow Z$ given a data probability measure $\mu$ on $X$, provided that $\mu$ can be written as the pushforward $G_X(\cdot, x_0)_\# \nu$ of some probability measure $\nu$ on $G$ by the function $G_X(\cdot, x_0)$ for some base point $x_0$. More formally,
\begin{equation}
\begin{aligned}
\mu(A) &= G_X (\cdot, x_0)_\# \nu(A) \\
&= \nu\left(\left\{ g\in G \ | \  G_X(g, x_0)\in A\right\}\right),
\end{aligned}
\end{equation}
for Borel subsets $A \subset X$. Note that this is only possible if the action $G_X$ is \emph{transitive}.

For example, the situation of a dataset with $N$ datapoints $\{x_n\}_{n=1}^N = \{ g_n\cdot x_0 \}_{n=1}^N$ corresponds to the case in which $\nu$ and $\mu$ are empirical measures on the group $G$ and data space $X$, respectively:
\begin{equation}\label{eq:empirical_measures}
\nu := \frac{1}{N} \sum_{i=1}^N \delta_{g_i}, \qquad \mu := \frac{1}{N} \sum_{i=1}^N \delta_{x_i}.
\end{equation}

We define the metric \metricname{} for an encoding $h$ and a measure $\mu$ as
\begin{multline}
    \label{eq:metric}
    \mathcal{D}_\mathrm{LSBD} := \\
    \inf_{\rho \in \mathcal{P}(G, Z)}
    \int_G \left\| \rho(g)^{-1} \cdot h(g \cdot x_0) - M_{\rho,h,x_0} \right\|_{\rho, h, \mu}^2 d \nu(g), \\
    \mbox{with} \; M_{\rho,h,x_0} = \int_G \rho(g')^{-1}\cdot h(g'\cdot x_0) d\nu(g'),
\end{multline}
where the norm $\| \cdot \|_{\rho, h, \mu}$ is a Hilbert-space norm depending on the representation $\rho$, the encoding map $h:X\rightarrow Z$, and the data measure $\mu$. More details of this norm can be found in Appendix~\ref{app:inner}. Moreover, $\mathcal{P}(G,Z)$ denotes the set of \emph{linearly disentangled representations} of $G$ in $Z$. Lower values of \metricname{} indicate better disentanglement, zero being optimal.

\subsection{Practical Computation of \metricname{}}
There are two main challenges for computing the metric of Equation~\eqref{eq:metric}. First, to calculate the integrals in the formula, all possible datapoints that can be expressed as $g\cdot x_0$ with $g\in G = G_1\times\cdots\times G_K$ must be available. Second, the infimum of the integrals over all possible linearly disentangled representations must be estimated. This requires finding the possible invariant subspaces $Z = Z_1\oplus\cdots \oplus Z_K$ induced by the encoding $h$ over which the group representations are disentangled.

We present a practical implementation of an upper bound to \metricname{} for an encoding function $h$ given a dataset $\mathcal{X}$ generated by some known group transformations. This approximation of \metricname{} is designed for a group decomposition $G = G_1\times\cdots\times G_K$ where each $G_k=\mathrm{SO}(D_k)$  with $k \in \{1,\ldots, K\}$ the group of rotations in $D_k$ dimensions. This implementation approximates the integrals of Equation~\eqref{eq:metric} by using the empirical distribution of $\mathcal{X}$. The invariant subspaces of $Z$ to the subgroup actions are found by applying a suitable change of basis. In the new basis, the disentangled group representations are expressed in a parametric form whose parameters are optimized to find the tightest bound to \metricname{}. See Fig.~\ref{fig:diagram_quantifying2} for an intuitive description of the process.

Assume there is a dataset $\mathcal{X}$ that can be modeled in terms of the group decomposition $G = G_1\times\cdots G_k$. For each $G_k$ subgroup there is a set of known group elements $\mathcal{G}_k\subseteq{G}_k$ uniformly sampled such that the dataset is described in terms of all elements in $\mathcal{G} = \mathcal{G}_1\times\cdots\times\mathcal{G}_K$ and a base point $x_0$ as $ \mathcal{X} = \left\{(g_1,\ldots, g_K)\cdot x_0 \middle| g_k\in\mathcal{G}_k,\;  k\in\{1,\ldots, K\}\right\}.$

For each subgroup $G_k$ we construct a set of encoded data $\mathcal{Z}_k\subseteq Z$ whose variability should only depend on the action of $G_k$. The set $\mathcal{Z}_k$ is given by $\mathcal{Z}_k = \left\{z_k(g_1,\ldots, g_K)\middle|  g_j\in \mathcal{G}_j\;,  j\in\{1,\ldots, K\}\right\}$, in which 
\begin{multline}
     z_k(g_1,\ldots, g_K) = h((g_1, \ldots, g_K)\cdot x_0) \\ -\frac{1}{|\mathcal{G}_k|}\sum_{g'\in \mathcal{G}_k}h((g_1,\ldots ,g_{k-1}, g',g_{k+1},\ldots, g_K)\cdot x_0).
\end{multline}

Similar to \citet{Cohen2014}, we find a suitable change of  basis that exposes the invariant subspace $Z_k$ corresponding to the $k$-th subgroup $G_k$. The new basis is obtained from the eigenvectors resulting from applying Principal Component Analysis (PCA) to $\mathcal{Z}_k$. Each element in $\mathcal{Z}_k$ is projected into the first $D_k$ eigenvectors. The new set is denoted as $\mathcal{Z}'_k\subseteq\mathbb{R}^{D_k}$ with elements $z'_k(g_1,\ldots, g_K )\subseteq \mathbb{R}^{D_k}$ that are the projected versions of $z_k(g_1,\ldots, g_K)$.

\citet{Quessard2020} describe how one could parameterize the subgroup representations of $SO(D_k)$ for arbitrary $D_k$ but here we will focus on $G_{k} = SO(2)$. In this case, we can parameterize each subgroup representation in terms of a single integer parameter $\omega\in\mathbb{Z}$ as $\rho_{k,\omega}(g_k)$ corresponding to a $2\times2$ rotation matrix whose angle of rotation is $\omega$ multiplied by the known angle associated to the  group element $g_k\in G_k = \mathrm{SO}(2)$. For this subgroup we can approximate the $M_{\rho,h,x_0}$ in Equation~\eqref{eq:metric} as $M_{k,\omega}$ given by

\begin{equation}
\begin{aligned}
   M_{k,\omega} =  \frac{1}{|\mathcal{G}|} \sum_{(g_1,\ldots, g_K)\in \mathcal{G}}\rho_{k,\omega}(g_k^{-1})\cdot z'(g_1,\ldots, g_K). 
\end{aligned}
\end{equation}
Similar to Equation~\eqref{eq:metric} we would like to find the optimal $\rho_{k,\omega}$ that minimizes the integral over the group representations. We can define a parameter search space $\Omega\subseteq \mathbb{Z}$, e.g. $\Omega = [-10,10]$ for finding the optimal $\omega\in\Omega$ that minimizes the dispersion, this is expressed in the following equation
\begin{multline}
 \mathcal{D}_{\mathrm{LSBD}}^{(k)} = \\ \min_{\omega\in\Omega}\frac{1}{|\mathcal{G}|}\sum_{(g_1,.., g_K)\in \mathcal{G} }\|\rho_{k,\omega}(g_k^{-1})\cdot z'(g_1,.., g_K)- M_{k,\omega}\|^2 .
\end{multline}

Each $\mathcal{D}_{\mathrm{LSBD}}^{(k)}$ measures the degree of equivariance of the projected embeddings for each $k$-th subgroup corresponding to the best fitting group representation. The upper bound to the metric is finally obtained by averaging across all subgroups: $\mathcal{D}_{\mathrm{LSBD}}\leq \frac{1}{K}\sum_{k=1}^K \mathcal{D}_{\mathrm{LSBD}}^{(k)}$.

Our practical implementation of \metricname{} is for $\mathrm{SO}(2)$ subgroups, however the procedure can in principle be extended to other subgroups as well. A practical implementation of the metric requires (i) identifying the subspaces invariant to a subgroup and (ii) identifying a parametric representation of the subgroup that can be fitted to the subspace data representations. In cases where the exact form of the subgroup is unknown, an option is to use the method by \citet{Pfau2020} to factorize the submanifolds associated with different generative factors.

\begin{figure*}[ht!]
\centering
\begin{minipage}{0.33\linewidth}
  \centering
  \includegraphics[width=\linewidth]{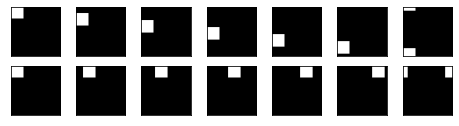}
  \subcaption{Square}
\end{minipage}
\centering
\begin{minipage}{0.33\linewidth}
  \centering
  \includegraphics[width=\linewidth]{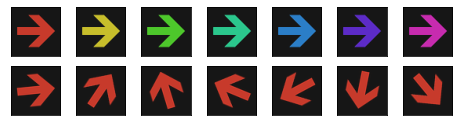}
  \subcaption{Arrow}
\end{minipage}
\centering
\begin{minipage}{0.33\linewidth}
  \centering
  \includegraphics[width=\linewidth]{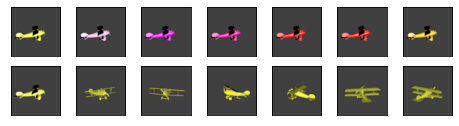}
  \subcaption{Airplane}
\end{minipage}
\begin{minipage}{0.33\linewidth}
  \centering
  \includegraphics[width=\linewidth]{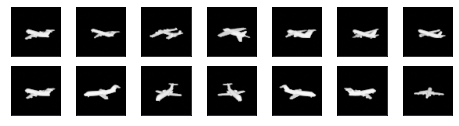}
  \subcaption{ModelNet40}
\end{minipage}
\begin{minipage}{0.33\linewidth}
  \centering
  \includegraphics[width=\linewidth]{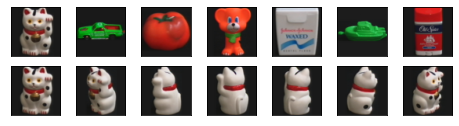}
  \subcaption{COIL-100}
\end{minipage}
  \caption{Example images from each of the datasets used.  Each row shows different examples from a single factor changing.}
  \label{fig:dataset_example}
\end{figure*}

\section{Learning LSBD: \methodname{}}
\label{sec:lsbd-vae}

In this section we present \methodname{}, a semi-supervised VAE-based method to learn LSBD representations. The main idea is to train an unsupervised Variational Autoencoder (VAE) \citep{Kingma2014, rezende2014stochastic} with a suitable latent space topology, and use our metric as an additional loss term for batches of transformation-labeled data.

\paragraph{Assumptions}
\methodname{} requires some knowledge about the group structure $G$ that is to be disentangled. Concretely, the group and its decomposition $G=G_1\times\ldots\times G_K$ should be known, as well as a suitable \emph{linearly disentangled} group representation $\rho: G \rightarrow \mathrm{GL}(Z)$ and a latent space $Z=Z_1\oplus\ldots\oplus Z_K$. Moreover, we assume there exists an embedded submanifold $Z_G \subseteq Z$ such that the action of $G$ on $Z$ restricted to $Z_G$ is \emph{regular}, and $Z_G$ is invariant under the action. Only $Z_G$ will then be used as the codomain for the encoding map, $h:X\rightarrow Z_G$.

We demonstrate the assumptions above for the common group structure $G = \mathrm{SO}(2)\times \mathrm{SO}(2)$. For the group representation $\rho=\rho_1\oplus\rho_2$, with $Z=\mathbb{R}^2\oplus\mathbb{R}^2$, we can use rotation matrices in $\mathbb{R}^2$ for $\rho_1$ and $\rho_2$. We can then use 1-spheres $S^1=\{z\in\mathbb{R}^2: \|z\|=1\}$ for the embedded submanifold: $Z_G = S^1 \times S^1$. In this case, the action of $G$ on $Z$ restricted to $Z_G$ is indeed \emph{regular}, and $Z_G$ is invariant under the action.

Requiring the group structure $G$ to be known is a relatively strong assumption, which limits the practical applicability of our method. However, a group structure can often be given as expert knowledge, like the presence of cyclic factors such as rotation, or in situations where transformations between observed data can easily be acquired such as in reinforcement learning.

\paragraph{Unsupervised Learning on Latent Manifold}
To learn encodings only on the latent manifold $Z_G$, we use a Diffusion Variational Autoencoder ($\Delta$VAE) \citep{Rey2019}. $\Delta$VAEs can use any closed Riemannian manifold embedded in a Euclidean space as a latent space (or latent manifold), provided that a certain \emph{projection function} from the Euclidean embedding space into the latent manifold is known and the \emph{scalar curvature} of the manifold is available. The $\Delta$VAE uses a parametric family of posterior approximates obtained from a diffusion process over the latent manifold. To estimate the intractable terms of the negative ELBO, the reparameterization trick is implemented via a random walk.

In the case of $S^1$ as a latent (sub)manifold, we consider $\mathbb{R}^2$ as the Euclidean embedding space, and the projection function\footnote{This projection function is not defined for $z=\mathbf{0}$, but this value does not occur in practice.} $\Pi: \mathbb{R}^2 \rightarrow S^1$ normalizes points in the embedding space: $\Pi(z) = z/|z|$. The scalar curvature of $S^1$ is $0$.

\paragraph{Semi-Supervised Learning with Transformation Labels}
\citet{Caselles-Dupre2019} proved that LSBD representations cannot be inferred from a training set of unlabeled observations, but that access to the transformations between data points is needed. They therefore use a training set of observation pairs with a given transformation between them.

However, we posit that only a limited amount of supervision is sufficient. Since obtaining supervision on transformations is typically more expensive than obtaining unsupervised observations, it is desirable to limit the amount of supervision needed.

\begin{figure}[ht!]
    \includegraphics[width=\linewidth]{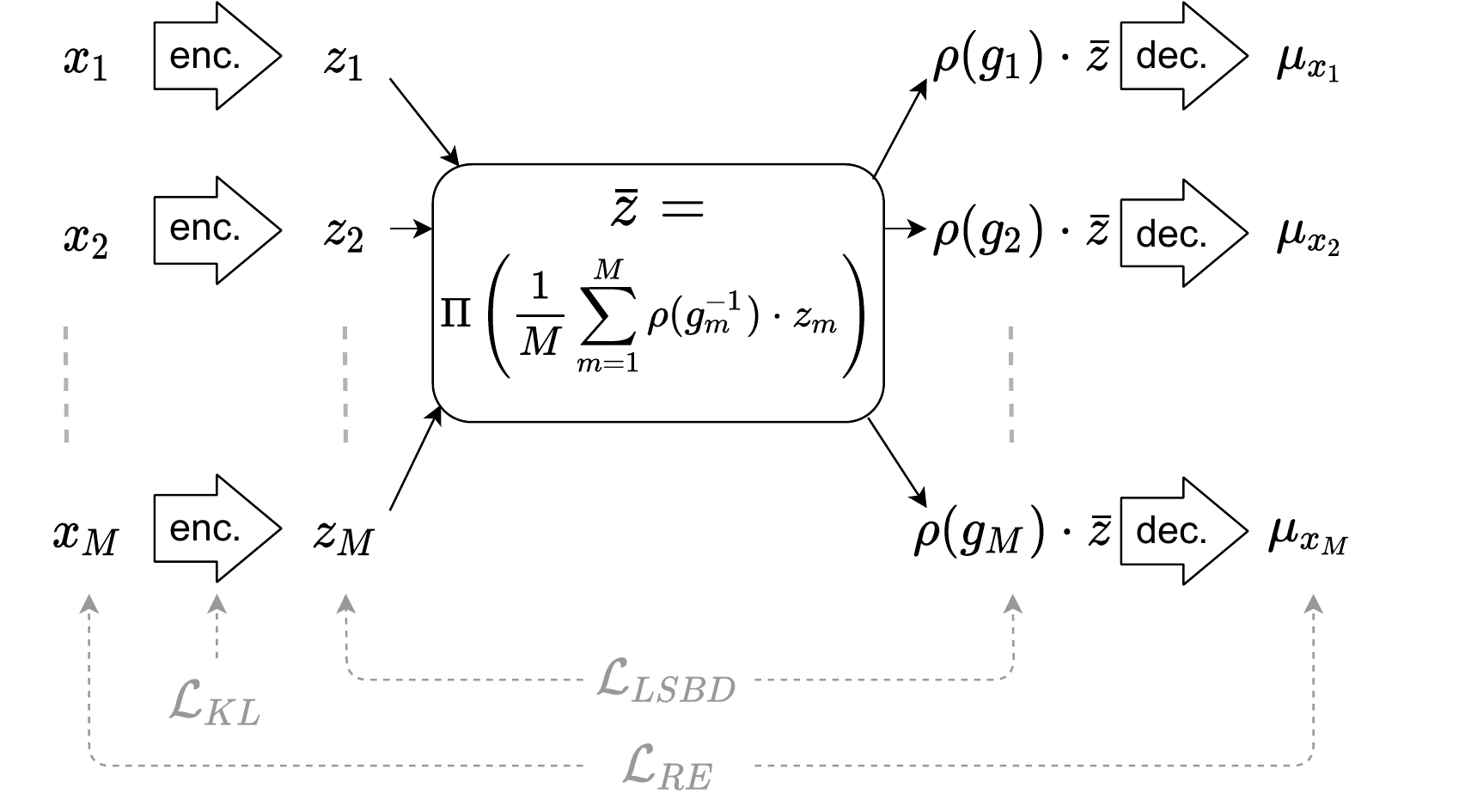}
    \centering
    \caption{Overview of the supervised part of \methodname{}.}
    \label{fig:method}
\end{figure}

Therefore, we augment the unsupervised $\Delta$VAE with a supervised method that makes use of transformation-labeled batches, i.e. batches $\{x_m\}_{m=1}^M$ such that $x_m=g_m\cdot x_1$ for $m=2,\ldots,M$, where the transformations $g_m$ (and thus their group representations $\rho(g_m)$) are known and are referred to as \emph{transformation labels}. The simplified version of the metric from Equation~\eqref{eq:simple_metric} can then be used for each batch as an additional loss term (with $x_0=x_1$), as it is differentiable under the assumptions described above (using the Euclidean norm).

We make a small adjustment to Equation~\eqref{eq:simple_metric} for the purpose of our method, since the mean computed there does not typically lie on the latent manifold $Z_G$. Thus, we use the projection $\Pi$ from the $\Delta$VAE to project the mean onto $Z_G$. Writing the encodings as $z_m := h(x_m)$, the additional loss term for a transformation-labeled batch $\{x_m\}_{m=1}^M$ becomes
\begin{multline}
    \mathcal{L}_{LSBD} = \\
    \frac{1}{M}\sum_{m=1}^M \left\| \rho(g_m^{-1})\cdot z_m - \Pi\left( \frac{1}{M}\sum_{m=1}^M \rho(g_m^{-1})\cdot z_m \right) \right\|^2,
\end{multline}
where $g_1 = e$, the group identity.

Moreover, instead of feeding the encodings $z_m$ to the decoder, we use $\rho(g_m)\cdot\overline{z}$, where $\overline{z}=\Pi\left( \frac{1}{M}\sum_{m=1}^M \rho(g_m^{-1})\cdot z_m \right)$. This encourages the decoder to follow the required group structure. This only affects the reconstruction loss component of the $\Delta$VAE.

Fig.~\ref{fig:method} illustrates the supervised part of our method for a transformation-labeled batch $\{x_m\}_{m=1}^M$. The loss function is the regular ELBO (but with adjusted decoder input as described above) as used in $\Delta$VAE plus an additional term $\gamma\cdot\mathcal{L}_{LSBD}$, where $\gamma$ is a weight hyperparameter to control the influence of the supervised loss component. By alternating unsupervised and supervised training (using the same encoder and decoder), we have a method that makes use of both unlabeled and transformation-labeled observations.

\begin{figure*}[ht!]
\centering
\begin{minipage}{0.62\linewidth}
  \centering
  \includegraphics[width=\linewidth]{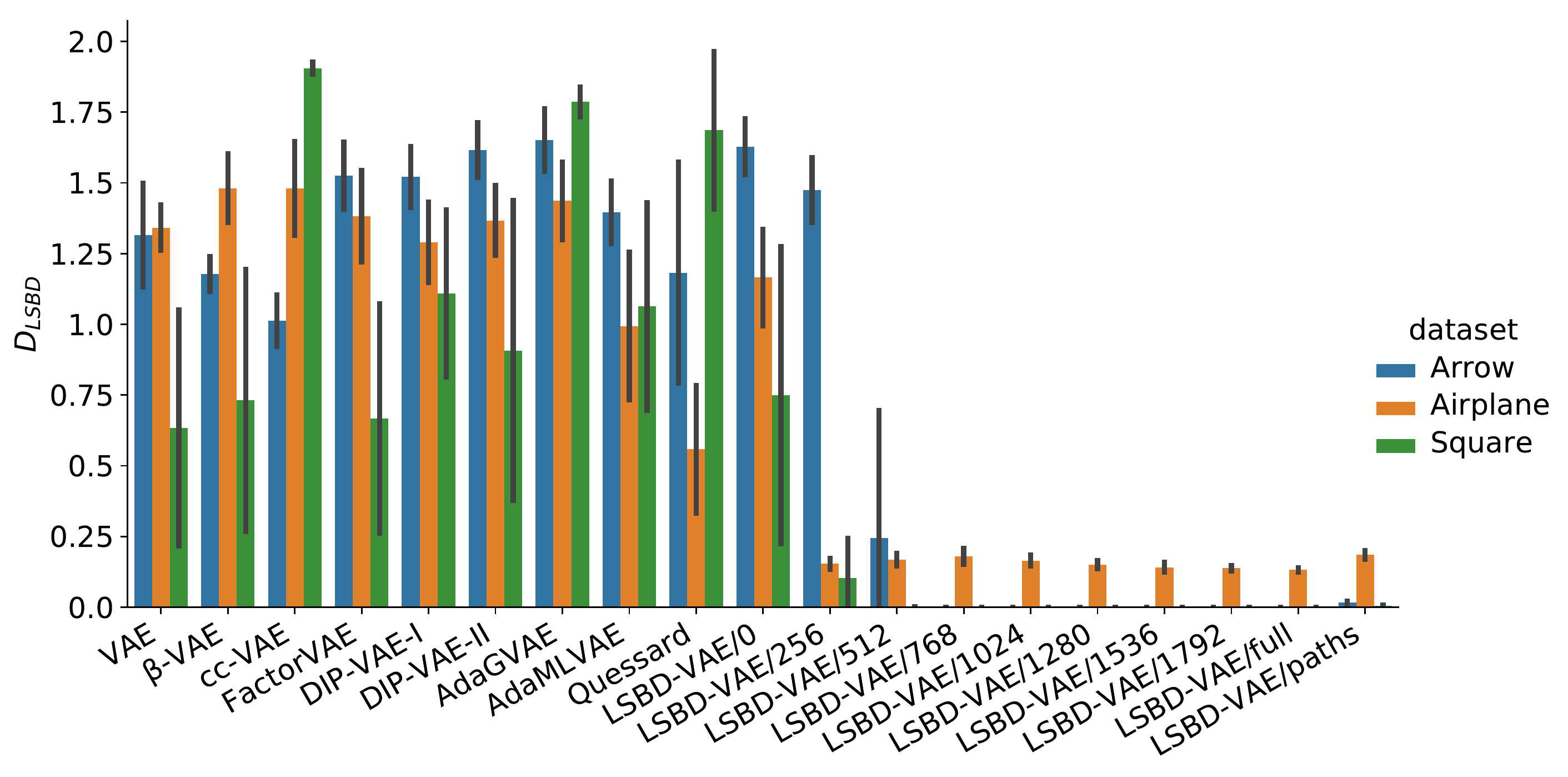}
  \subcaption{Datasets with  $\mathrm{SO}(2)\times \mathrm{SO}(2)$ symmetries}
\end{minipage}
\centering
\begin{minipage}{0.36\linewidth}
  \centering
  \includegraphics[width=\linewidth]{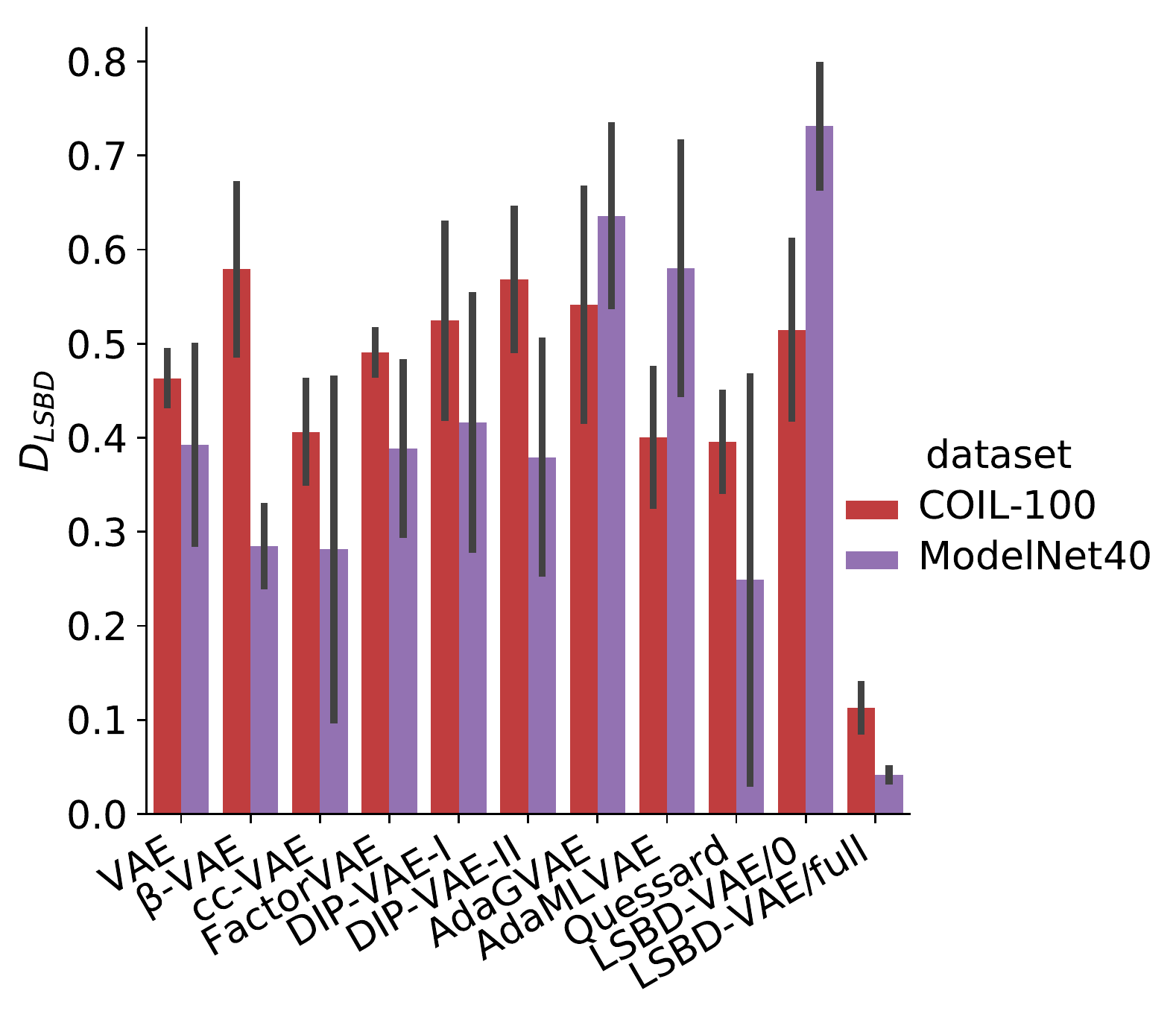}
  \subcaption{Datasets with $\mathrm{SO}(2)$ and non-symmetric variation}
\end{minipage}
  \caption{\metricname{} scores for all methods on all datasets}
  \label{fig:lsbd_scores}
\end{figure*}

\section{Experimental Setup}
\label{sec:experimental_setup}

\paragraph{Data}
We evaluate the disentanglement of several models on three different image datasets (Square, Arrow, and Airplane) with a known group decomposition $G=\mathrm{SO}(2)\times\mathrm{SO}(2)$ describing the underlying transformations. For each subgroup a fixed number of $|\mathcal{G}_k| = 64$ with $k\in \{1,2\}$ transformations is selected. The datasets exemplify different group actions of $\mathrm{SO}(2)$: periodic translations, in-plane rotations, out-of-plane rotations, and periodic hue-shifts.

In real settings, not all variability in the data can be modelled by the actions of a group. Therefore, we also evaluate the same models on two datasets ModelNet40 \citep{Wu2014}  and COIL-100 \citep{nene1996columbia} that consist of images from various objects (i.e. non-symmetric variation) under known out-of-plane rotations ($\mathrm{SO}(2)$ symmetries). In many settings it is easy to obtain labels for such rotations, e.g. when the camera or object angle is controlled by an agent. See Fig.~\ref{fig:dataset_example} for examples of the datasets. For more details, see Appendix~\ref{app:datasets}.

Note that we do not evaluate our \methodname{} method and \metricname{} metric on traditional disentanglement datasets as evaluated by \citet{locatello2019challenging}, since these datasets lack a clear underlying group structure. However, our results on the ModelNet40 and COIL-100 datasets show that our method can disentangle properties with a group structure from properties without such a structure.

\paragraph{\methodname{} with Semi-Supervised Labelled Pairs}
For the Square, Arrow, and Airplane datasets we test \methodname{} with transformation-labeled batches of size $M=2$. More specifically, for each experiment we randomly select $L$ disjoint pairs of data points, and label the transformation between the data points in each pair. We vary the number of labeled pairs $L$ from 0 (corresponding to a $\Delta$VAE) to $N/2$ (in which case each data point is involved in exactly one labeled pair). We set the weight $\gamma$ of the supervised loss component to $\gamma=100$ for all experiments. We choose $M=2$ for our experiments since it is the most limited setting for \methodname{}. Higher values of $M$ would provide stronger supervision, so successful results with $M=2$ imply that good results can also be achieved for higher values of $M$ (but not necessarily vice versa).

For the COIL-100 and ModelNet40 datasets, we train \methodname{} on batches containing images of one particular object from all different angles (72 and 64 for COIL-100 and ModelNet40, respectively). Each batch is labelled with transformations $(g_1, e), \ldots, (g_M, e)$, where $g_m$ represent rotations, and the unit transformation $e$ indicates that the object is unchanged. To represent the rotations we use a $S^1$ latent space as in $\Delta$VAE, whereas for the object identity we use a 5-dimensional Euclidean space with standard Gaussian prior as in regular VAEs. LSBD is measured as the disentanglement of rotations in the latent space. For these experiments we used $\gamma=1$.

\paragraph{\methodname{} with Paths of Consecutive Observations}
It is often cheap to obtain transformation labels in settings where we can apply simple transformations and observe its effect, such as an agent navigating its environment. By registering actions (e.g. rotate left over a given angle) and the resulting observations, we can construct a path of consecutive views with known in-between transformations. We can then use these paths to train a \methodname{}.

For the datasets with $G=G_1 \times G_2 = \mathrm{SO}(2)\times\mathrm{SO}(2)$ (Square, Arrow, Airplane), we generate random paths by consecutively applying one randomly chosen transformation from $\{g_1, g_1^{-1}, g_2, g_2^{-1}\}$ where $g_k \in G_k$ for $k\in \{1, 2\}$, starting from randomly chosen observations. In our experiments, we generate 50 paths of length 100, and $g_k$ corresponds to an SO(2) transformation corresponding to an angle of $\frac{3}{64} 2\pi$ radians. Example paths can be found in Fig.~\ref{fig:path_examples} in the Appendix.

For the COIL-100 and ModelNet40 datasets there is only one group to disentangle. Therefore, similar random walks are not very meaningful here, and we do not evaluate them for these datasets.


\paragraph{Other Disentanglement Methods}
We furthermore test a number of known disentanglement methods for comparison, including traditional disentanglement methods as well as methods focusing on LSBD. In particular, we use \texttt{disentanglement\_lib} \citep{locatello2019challenging} to train a regular VAE \citep{Kingma2014, rezende2014stochastic}, $\beta$-VAE \citep{higgins2016betavae}, CC-VAE \citep{burgess2018understanding}, FactorVAE \citep{kim2018factorvae}, and DIP-VAE-I/II \citep{kumar2017dip-vae}. We also include two weakly-supervised models, AdaGVAE and AdaMLVAE \citep{Locatello2020}, which are trained on pairs of data with few changing factors, to test whether this kind of supervision is helpful for LSBD. Furthermore we evaluate the method from \citet{Quessard2020} that focuses on LSBD. We also tested ForwardVAE \citep{Caselles-Dupre2019}, but show only limited results since we were not able to reproduce any reasonable results for our datasets.

Most of these methods have no notion of an underlying group structure, and thus do not give a fully fair comparison with our \methodname{} method. However, we emphasize that the main goal of our experiments is to investigate properties of disentangled representations from both the traditional and the LSBD perspective.

\paragraph{Disentanglement Metrics}
We use encodings from all methods to evaluate \metricname{}, as well as common traditional disentanglement metrics from \texttt{disentanglement\_lib}: Beta \citep{higgins2016betavae}, Factor \citep{kim2018factorvae}, SAP \citep{kumar2017dip-vae}, DCI Disentanglement \citep{eastwood2018framework}, Mutual Information Gap (MIG) \citep{chen2018betatcvae}, and Modularity (MOD) \citep{ridgeway2018learning}.

\paragraph{Further Details}
More information about the architectures, epochs and hyperparameters can be found in Appendix~\ref{app:exp_settings}. For the traditional disentanglement methods trained on Square, Arrow and Airplane datasets the latent spaces have 4 dimensions, since these are the minimum number of dimensions necessary to learn LSBD representations for an underlying $\mathrm{SO}(2)\times\mathrm{SO}(2)$ symmetry group, see \citep{higgins2018towards, Caselles-Dupre2019}. For COIL-100 and ModelNet40 we use latent spaces with 7 dimensions for a fair comparison with the LSBD-VAE method.

\begin{figure*}[ht!]
\centering
\begin{minipage}{0.49\linewidth}
  \centering
  \includegraphics[width=\linewidth]{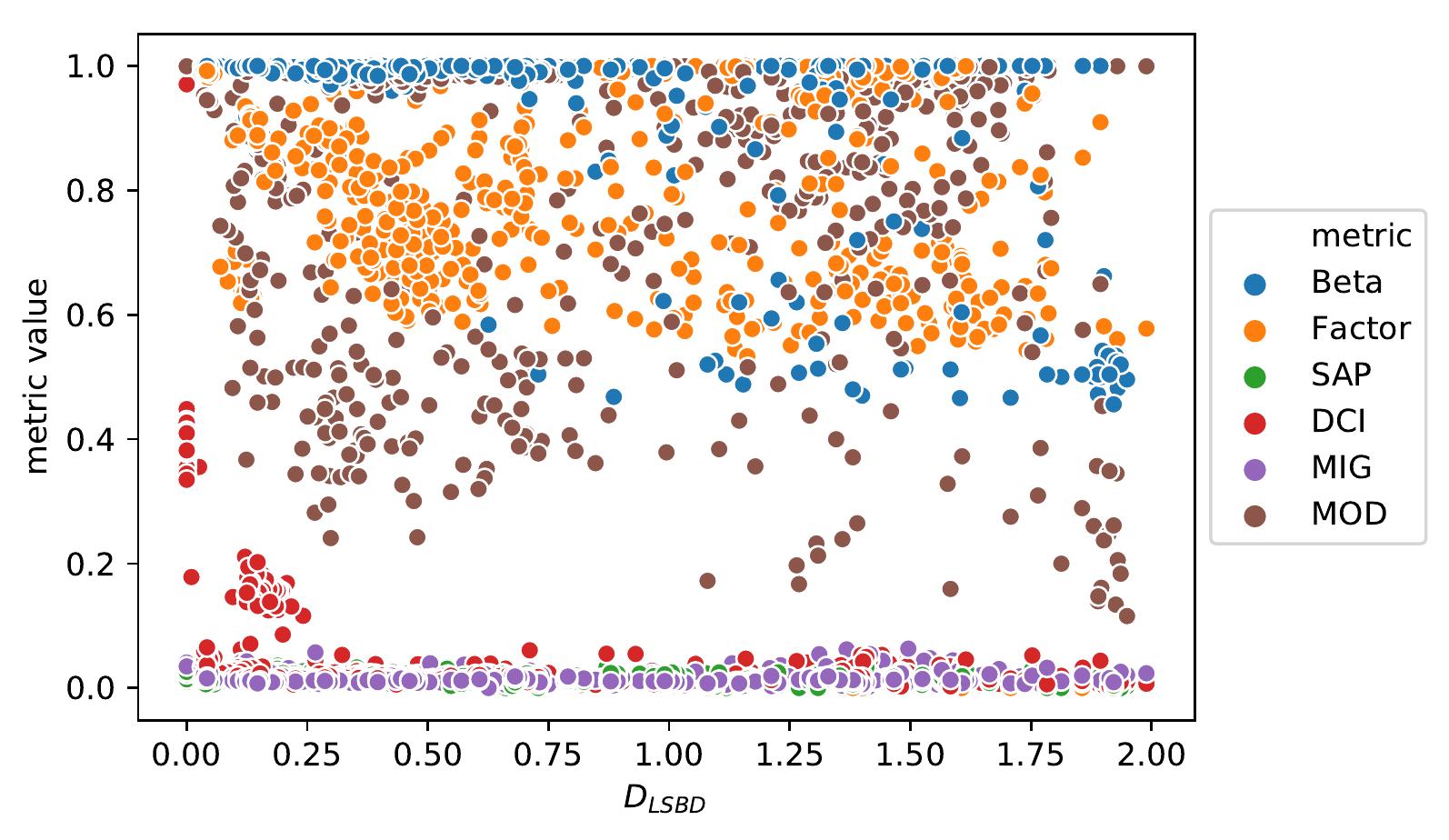}
  \subcaption{Scatter plot}
\end{minipage}
\centering
\begin{minipage}{0.49\linewidth}
  \centering
  \includegraphics[width=\linewidth]{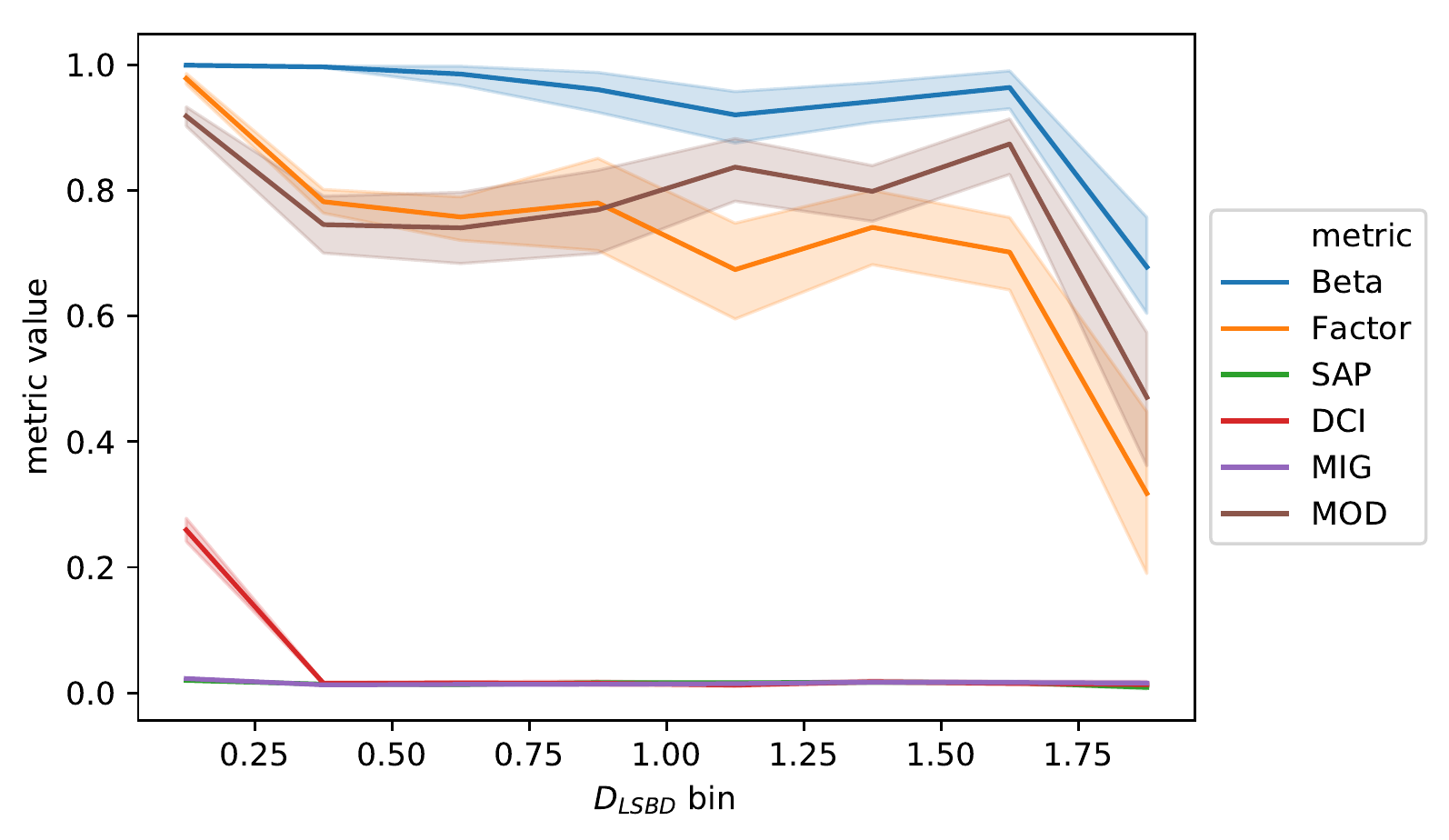}
  \subcaption{Binned by \metricname{} value}
\end{minipage}
  \caption{Comparing \metricname{} to previous disentanglement metrics}
  \label{fig:dis_metrics}
\end{figure*}

\section{Results: Evaluating LSBD with \metricname{}}
\label{sec:results}
We now highlight three key observations from our experimental results. In particular, we differentiate between the methods ({\sc VAE, $\beta$-VAE, cc-VAE, Factor, DIP-I, DIP-II}) and metrics ({\sc Beta, Factor, SAP, DCI, MIG, MOD}) that approach disentanglement in the \emph{traditional} sense, and methods ({\sc $\Delta$VAE, Quessard, \methodname{}}) and metric (\metricname{}) that focus specifically on LSBD. The full quantitative results can be found in Appendix~\ref{app:full_results}. Further qualitative results can be found in Appendix~\ref{app:qualitative_results}.

\subsection{Traditional Disentanglement Methods Don't Learn LSBD Representations}
\label{subsec:results_no_lsbd}
Fig.~\ref{fig:lsbd_scores} summarizes the \metricname{} scores (lower is better) for all methods on all datasets. Bars show the mean scores over 10 runs for each method, the vertical lines represent standard deviations. \methodname{}/$L$ indicates our method trained on $L$ labelled pairs (\methodname{}/0 corresponds to the unsupervised $\Delta$VAE), \methodname{}/full indicates our method where all images are involved in exactly one labelled pair. and \methodname{}/paths indicates our method trained with paths of consecutive observations. Note that \methodname{} obtained very good scores (near 0) on the Arrow and Square datasets, hence the missing bars.

None of the traditional disentanglement methods achieve good \metricname{} scores, even if they score well on other traditional disentanglement metrics. This implies that LSBD isn't achieved by traditional methods. Moreover, from the full results in Appendix~\ref{app:full_results} we see that the traditional methods on these datasets do not achieve good scores on all traditional metrics. In particular, SAP, DCI, and MIG scores are low. We believe this is a result of the cyclic nature of the symmetries underlying our datasets, further emphasizing the need for disentanglement methods that can capture such symmetries.

The SAP and MIG scores measure to what extent generative factors are disentangled into a single latent dimension. However, since the factors in our dataset are inherently cyclic due to their symmetry structure, they cannot be properly represented in a single latent dimension, as shown by \citet{Rey2019}. Instead, at least two dimensions are needed to continuously represent each cyclic factor in our data. A similar conclusion was made by \citet{Caselles-Dupre2019} and \citet{painter2020linear}.

DCI disentanglement measures whether a latent dimension captures at most one generative factor. This is accomplished by measuring the importance of each latent dimension in predicting the true generative factor using boosted trees. However, since the generative factors are cyclic, the performance of the boosted tree classifiers is far from optimal, thus providing more importance to several dimensions in predicting the generative factors and giving overall lower DCI scores.

\subsection{\methodname{} and other LSBD Methods \emph{Can} Learn LSBD Representations with Limited Supervision on Transformations}

From Fig.~\ref{fig:lsbd_scores} we observe that methods focusing specifically on LSBD can score higher on \metricname{}, showing that they are indeed more suitable to learn LSBD representations. In particular, \methodname{} got very good \metricname{} scores for all datasets. Moreover, our experiments on the Arrow, Airplane, and Square datasets also show that only limited supervision suffices to obtain good \metricname{} scores with low variability, either with few transformation-labelled pairs or with paths of consecutive observations that are easy to obtain in agent-environment settings.

We only partially managed to reproduce the results from \citet{Quessard2020} on our datasets. Their method scored fairly well on the Airplane, ModelNet40, and COIL-100 datasets, but did not do well on the Square and Arrow dataset in our experiments. 

Furthermore, we tested ForwardVAE by \citet{Caselles-Dupre2019}, but we could not produce any reasonable results on our datasets. Therefore, we do not include scores for this method. We did manage to reproduce ForwardVAE's results on the Flatland dataset used in the original paper, for which we computed a mean \metricname{} score of 0.012 with standard deviation 0.001 over 10 runs, confirming that ForwardVAE indeed learns LSBD representations for Flatland.

\subsection{LSBD Representations Also Satisfy Previous Disentanglement Notions}
Our results also indicate that LSBD captures various desirable properties that are expressed by traditional disentanglement metrics. In Fig.~\ref{fig:dis_metrics} we compare \metricname{} scores with scores for previous disentanglement metrics. Note that for \metricname{} lower is better, whereas for all other metrics higher is better. As we noted before, good scores on traditional disentanglement metrics don't necessarily imply good \metricname{} scores. Conversely however, methods that score well on \metricname{} also score well on many traditional disentanglement metrics, often even outperforming the traditional methods. In particular, from the full results (see Appendix~\ref{app:full_results}) we see that \methodname{} matches or outperforms the traditional methods on the {\sc Beta, Factor} and MOD metrics, and achieves much better scores for the DCI metric where traditional methods scored poorly.

The MIG and SAP scores are still low for methods focusing on LSBD. This is expected however, as explained earlier in Section~\ref{subsec:results_no_lsbd}. This was also observed by \citet{painter2020linear} for different datasets.

\section{Conclusion}
We presented \metricname{}, a metric to quantify Linear Symmetry-Based Disentanglement (LSBD) as defined by \citet{higgins2018towards}. We used this metric formulation to motivate \methodname{}, a semi-supervised method to learn LSBD representations given some expert knowledge on the underlying group symmetries that are to be disentangled.

We used \metricname{} to evaluate various disentanglement methods, both traditional methods and recent methods that specifically focus on LSBD, and showed that \methodname{} can learn LSBD representations where traditional methods fail to do so. We also compared \metricname{} to traditional disentanglement metrics, showing that LSBD captures many of the same desirable properties that are expressed by existing disentanglement methods. Conversely, we also showed that traditional disentanglement methods and metrics do not usually achieve or measure LSBD.

Challenges that remain are expanding and testing \methodname{} and \metricname{} on different group structures, towards more practical applications, as well as focusing on the utility of LSBD representations for downstream tasks.

\section{Acknowledgements}
This work has received funding from the Electronic Component Systems for European Leadership Joint Undertaking under grant agreement No 737459 (project Productive4.0). This Joint Undertaking receives support from the European Union Horizon 2020 research and innovation program and Germany, Austria, France, Czech Republic, Netherlands, Belgium, Spain, Greece, Sweden, Italy, Ireland, Poland, Hungary, Portugal, Denmark, Finland, Luxembourg, Norway, Turkey.

This work has also received funding from the NWO-TTW Programme “Efficient Deep Learning” (EDL) P16-25.

\bibliography{mybibliography}
\bibliographystyle{icml2022}

\newpage
\appendix
\onecolumn

\section{Preliminaries: Group Theory}
\label{app:preliminaries}
In this appendix, we summarize some concepts from group theory that are important to understand the main text of the paper. Group theory provides a useful language to formalize the notion of symmetry transformations and their effects. For a more elaborate discussion we refer the reader to the book from \citet{Hall2015} on group theory.

\paragraph{Group}
A group is a non-empty set $G$ together with a binary operation $\circ: G\times G\rightarrow G$ that satisfies three properties:
\begin{enumerate}
    \item \textit{Associativity}: For all $f,g,h\in G$, it holds that $f\circ \left( g\circ h\right)  = \left(f\circ  g\right)\circ h $.
    \item \textit{Identity}: There exists a unique element $e\in G $ such that for all $g\in G$ it holds that $e\circ g = g\circ e = g$.
    \item \textit{Inverse}: For all $ g\in G$ there exists an element $g^{-1}\in G$ such that $g^{-1}\circ g = g \circ g^{-1} = e$.
\end{enumerate}

\paragraph{Direct product}
Let $G$ and $G'$ be two groups. The \emph{direct product}, denoted by $G\times G'$, is the group with elements $(g,g')\in G\times G'$ with $g\in G$ and $g'\in G'$, and the binary operation $\circ:G\times G'\rightarrow G\times G'$ such that $(g,g')\circ (h,h')= (g\circ h, g'\circ h')$.

\paragraph{Lie group}
A Lie group is a group where $G$ is a smooth manifold, this means it can be described in a local scale with a set of continuous parameters and that one can interpolate continuously between elements of $G$.

\paragraph{Group action}
Let $A$ be a set and $G$ a group. The \emph{group action} of $G$ on $A$ is a function \mbox{$G_A: G\times A\rightarrow A$} that has the properties \footnote{To avoid notational clutter, we write $G_A(g,a) = g\cdot a$ where the set $A$ on which $g\in G$ acts can be inferred from the context.}
\begin{enumerate}
    \item $G_A(e, a) = a$ for all $a\in A$
    \item $G_A(g,(G_A(g',a)) = G_A(g\circ g',a)$ for all $ g,g'\in G $ and $a\in A$
\end{enumerate}

\paragraph{Regular action}
The action of $G$ on $A$ is regular if for every pair of elements $a,a' \in A$ there exists a unique $g\in G$ such that $g\cdot a = a'$.

\paragraph{Group representation}
A \emph{group representation} of $G$ in the vector space $V$ is a function $\rho:G\rightarrow GL(V)$ (where $GL(V)$ is the general linear group on $V$) such that for all $ g,g'\in G$ $\rho(g\circ g') = \rho(g) \circ \rho(g')$ and $\rho(e) = \mathbb{I}_V$, where $\mathbb{I}_V$ is the identity matrix.

\paragraph{Direct sum of representations}
The direct sum of two representations $\rho_1:G\rightarrow GL(V)$ in $V$ and $\rho_2:G\rightarrow GL(V')$ in $V'$ is a group representation $\rho_1\oplus\rho_2:G\rightarrow GL(V\oplus  V')$ over the direct sum $V\oplus V'$, defined for $v\in V$ and $v'\in V'$ as:
\begin{equation}
    (\rho_1\oplus \rho_2)(g)\cdot(v,v') = \left(\rho_1(g)\cdot v, \rho_2(g)\cdot v'\right)
\end{equation}

\section{Linear Symmetry-Based Disentanglement: Definition with respect to World States}
\label{app:world_states}
\citet{higgins2018towards} provide a formal definition of linear disentanglement that connects symmetry transformations affecting the real world (from which data is generated) to the internal representations of a model. In the main text, we provide a definition from the perspective of a group action on the data directly, but the original definition considers an extra conceptual world state as well. Here, we describe the original setting in more detail, and explain why we choose a more direct and practical version of the definition.

The definition assumes the following setting. $W$ is the set of possible world states, with underlying symmetry transformations that are described by a group $G$ and its action $\cdot: G\times W\rightarrow W$ on $W$. In particular, $G$ can be decomposed as the direct product of $K$ groups $G = G_1\times\ldots\times G_K$. Data is obtained via an \emph{observation} function $b:W\rightarrow X$ that maps world states to observations in a \emph{data space} $X$. A model's internal representation of data is modeled with the \emph{encoding} function $h:X\rightarrow Z$ that maps data to the \emph{embedding space} $Z$. Together, the observation and the encoding constitute the model's internal representation of the real world $f:W\rightarrow Z$ with $f(w) = h\circ b(w)$. The definition for Linearly Symmetry-Based Disentangled (LSBD) representations then formalizes the requirement that a model's internal representation $f$ should reflect and disentangle the transformation properties of the real world, and that the transformation properties of the model's internal representations should be linear.

The original definition considers $G$ acting on $W$ and involves the model's internal representation $f:W\rightarrow Z$, but since we do not directly observe $W$ it is more practical to evaluate LSBD with respect to the encoding map $h:X\rightarrow Z$ instead. If the action of $G$ on $W$ is \emph{regular}\footnote{This assumption holds in most practical cases with a suitable description of $G$.} and the observation map $b:W\rightarrow X$ is \emph{injective}\footnote{This is typically the case, but if not it can be solved through active sensing, see~\citet{Soatto2011}.} though, we can instead define LSBD with respect to the action of $G$ on $X$ and the encoding map $h$, as shown in the main text.

\section{Inner Product}
\label{app:inner}

To describe the norm $\| \cdot \|_{\rho, h, \mu}$ used in the definition of \metricname{} we start with an arbitrary inner product $(\cdot, \cdot)$ on the linear latent space $Z$.
Assume that $\rho$ is linearly disentangled and accordingly splits in irreducible representations $\rho_k : G \to Z_k$ where $Z = Z_1 \oplus \cdots \oplus Z_K$ for some $K \in \mathbb{N}$.
We will define a new inner product $\langle \cdot, \cdot \rangle_{\rho, h, \mu}$ on $Z$ as follows.
First of all we declare $Z_k$ and $Z_m$ to be orthogonal with respect to $\langle \cdot, \cdot \rangle_{\rho, h, \mu}$ if $k \neq m$. We denote by $\pi_k$ the orthogonal projection on $Z_k$.

For $z, z' \in Z_i$, we set
\begin{equation}
\langle z, z' \rangle_{\rho, h, \mu} := \lambda_{k, h, \mu}^{-1} \int_{g \in G} (\rho(g)\cdot z, \rho(g)\cdot z' ) d \mathfrak{m}(g)
\end{equation}
where $\mathfrak{m}$ is the (bi-invariant) Haar measure normalized such that $\mathfrak{m}(G) = 1$ and set
\begin{equation}
\lambda_{k, h, \mu} := \int_X \int_G \|\pi_k(h(x))\|^2 d \mathfrak{m}(g) d\mu(x)
\end{equation}
if the integral on the right-hand side is strictly positive and otherwise we set $\lambda_k := 1$. This construction completely specifies the new inner product, and it has the following properties:
\begin{itemize}
    \item the subspaces $Z_k$ are mutually orthogonal,
    \item $\rho_k(g)$ is orthogonal on $Z_k$ for every $g \in G$, in other words $\rho_k$ maps to the orthogonal group on $Z_k$. Moreover, $\rho$ maps to the orthogonal group on $Z$. This follows directly from the bi-invariance of the Haar measure and the definition of $\langle\cdot, \cdot\rangle_{\rho, h, \mu}$.
    \item If $\pi_k$ is the orthogonal projection to $Z_k$, then 
    \begin{equation}
    \int_X \| \pi_k(h(x)) \|_{\rho, h, \mu}^2  d\mu(x) = 1
    \end{equation}
    if the integral on the left is strictly positive.
\end{itemize}

For an arbitrary pair $z, z'\in Z$ the inner product $\langle\cdot,\cdot \rangle_{\rho, h, \mu}$ is given by 
\begin{equation}
    \langle z, z'\rangle_{\rho, h, \mu} = \sum_{k = 1}^K \lambda_{k, h, \mu}^{-1}\int_{g\in G} (\rho(g)\cdot \pi_{k}(z), \rho(g)\cdot\pi_{k}(z'))d\mathfrak{m}(g)
\end{equation}

\section{Evaluation of Equivariance by \metricname{}}\label{app:equiv_m_lsbd}
We will now give an alternative expression for the disentanglement metric \metricname{}, since it will more visibly relate to the definition of equivariance. To avoid notational cluttering, in this section we will denote the norm $\|\cdot\|_{\rho, h, \mu}$ as $\| \cdot \|_*$.
Let $\rho\in \mathcal{P}(G,Z)$ be a linear disentangled representation of $G$ in $Z$. By expanding the inner product (or by using usual computation rules for expectations and variances), we first find that
\begin{equation}
\begin{split}
&\int_G \left\|\rho(g)^{-1}\cdot h(g\cdot x_0) - \int_G \rho(g')^{-1} \cdot h(g'\cdot x_0)d\nu(g') \right\|_*^2 d \nu(g)\\
&= \int_G \left\| \rho(g)^{-1}\cdot h(g\cdot x_0) \right\|_*^2 d\nu(g)- \left\|\int_G \rho(g)^{-1}\cdot h(g\cdot x_0) d\nu(g)\right\|_*^2\\
&= \frac{1}{2}\int_G \int_G \| \rho(g)^{-1}\cdot h(g\cdot x_0)- \rho(g')^{-1} \cdot h(g'\cdot x_0) \|_*^2 d \nu(g) d\nu(g').
\end{split}
\end{equation}
We now use that $\rho$ maps to the orthogonal group for $(\cdot, \cdot)_*$, so that we can write the same expression as
\begin{equation}
\frac{1}{2}\int_G \int_G \| \rho(g\circ g'^{-1})^{-1} \cdot h(((g\circ g'^{-1})\cdot g')\cdot x_0) - h(g'\cdot x_0) \|_*^2 d \nu(g) d\nu(g').
\end{equation}
This brings us to the alternative characterization of $\mathcal{D}_{\mathrm{LSBD}}$ as
\begin{equation}
\mathcal{D}_{\mathrm{LSBD}} = \inf_{\rho \in \mathcal{P}(G,Z)} \frac{1}{2}\int_G \int_G \|\rho(g\circ g'^{-1})^{-1} h(((g\circ  g'^{-1})\cdot g') \cdot x_0) - h(g'\cdot x_0)\|_{*}^2 d \nu(g) d \nu(g').
\end{equation}
In particular, if for every data point $x$ there is a unique group element $g_x$ such that $x = g_x\cdot x_0$, the disentanglement metric \metricname{} can also be written as
\begin{equation}
\inf_{\rho \in \mathcal{P}(G,Z)} \frac{1}{2}\int_G \int_X \|\rho(g\circ  g_x^{-1})^{-1} h((g\circ g_x^{-1})\cdot x) - h(x)\|_{*}^2 d \nu(g) d \mu(x), 
\end{equation}
in which the equivariance condition appears prominently. The condition becomes even more apparent if $\nu$ is in fact the Haar measure itself, in which case the metric equals
\begin{equation}
\inf_{\rho \in \mathcal{P}(G,Z)} \frac{1}{2}\int_{G} \int_X \|\rho(g)^{-1} \circ h(g\cdot x) - h(x)\|_*^2 d \mathfrak{m}(g) d\mu(x).
\end{equation}

\begin{figure*}[ht!]
\centering
\begin{minipage}{0.3\linewidth}
  \centering
  \includegraphics[width=\linewidth]{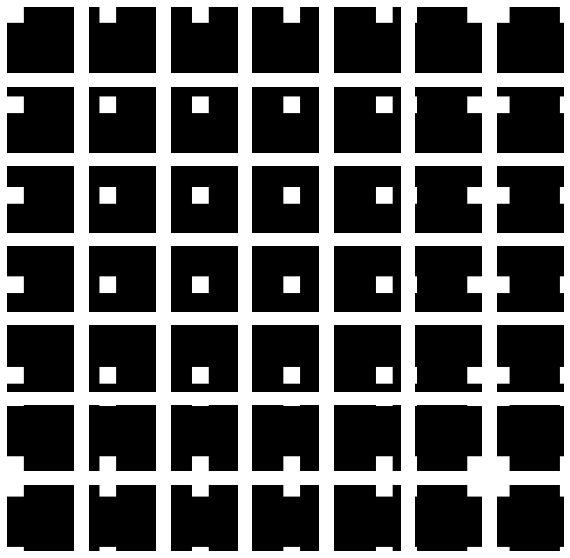}
  \subcaption{Square}
\end{minipage}
\centering
\begin{minipage}{0.3\linewidth}
  \centering
  \includegraphics[width=\linewidth]{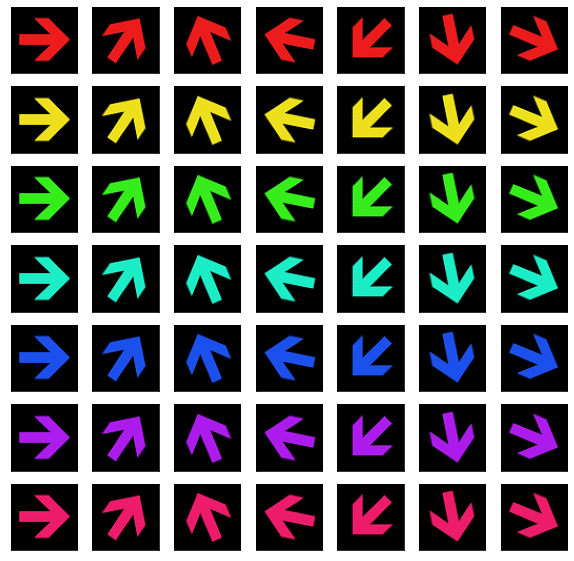}
  \subcaption{Arrow}
\end{minipage}
\centering
\begin{minipage}{0.3\linewidth}
  \centering
  \includegraphics[width=\linewidth]{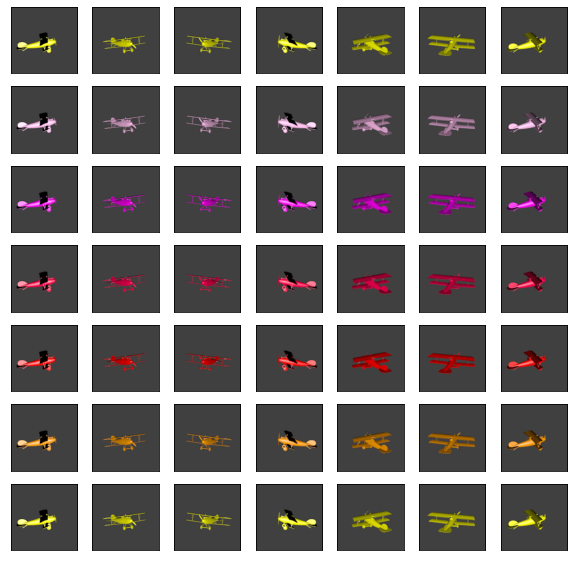}
  \subcaption{Airplane}
\end{minipage}

\begin{minipage}{0.3\linewidth}
  \centering
  \includegraphics[width=\linewidth]{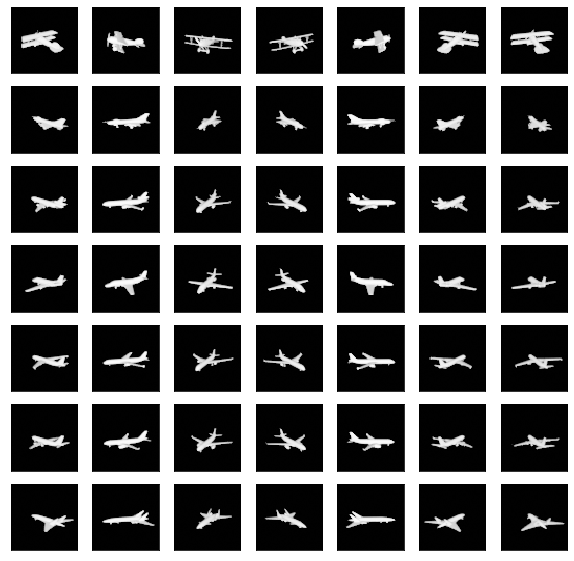}
  \subcaption{ModelNet40}
\end{minipage}
\begin{minipage}{0.3\linewidth}
  \centering
  \includegraphics[width=\linewidth]{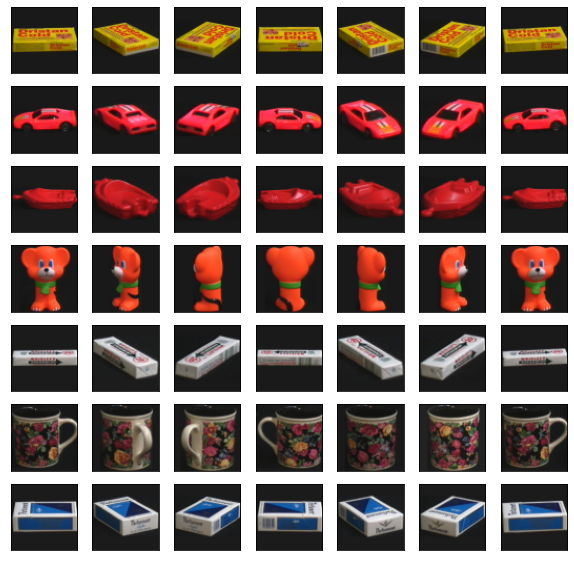}
  \subcaption{COIL-100}
\end{minipage}
  \caption{Example images from each of the datasets used. Each image corresponds to an example data point for a combination of two factors, e.g. color and orientation. The factors change horizontally and vertically. The boundaries for the Square, Arrow and Airplane dataset are periodic. For the ModelNet40 and COIL-100 dataset, the vertical direction represents different object instances and the horizontal direction represents the rotation of the corresponding object. }
  \label{fig:dataset_example2}
\end{figure*}

\section{Datasets}
\label{app:datasets}
All datasets contain $64 \times 64$ pixel images. The Square, Arrow and Airplane datasets have a known group decomposition $G = \mathrm{SO}(2) \times \mathrm{SO}(2)$ describing the underlying transformations. In these three datasets, for each subgroup a fixed number of $|\mathcal{G}_k| = 64$ with $k\in \{1,2\}$ transformations is selected. Each image is generated from a single initial data point upon which all possible group actions are applied, resulting in datasets with $|\mathcal{G}_1|\cdot |\mathcal{G}_2| = 4096$ images. The datasets exemplify different group actions of $\mathrm{SO}(2)$: periodic translations, in-plane rotations, out-of-plane rotations, and periodic hue-shifts, see Fig.~\ref{fig:dataset_example2}.

\begin{figure*}[ht!]
\centering
\begin{minipage}{\linewidth}
  \centering
  \includegraphics[width=\linewidth]{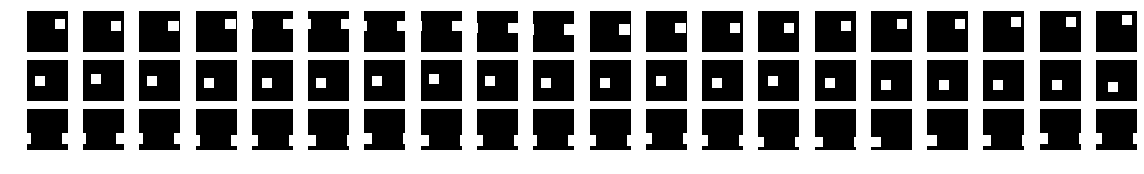}
  \subcaption{Square}
\end{minipage}
\centering
\begin{minipage}{\linewidth}
  \centering
  \includegraphics[width=\linewidth]{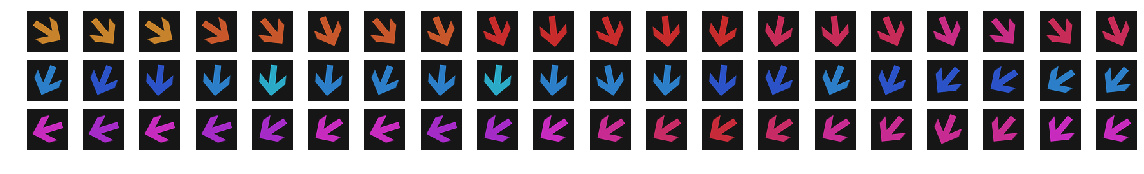}
  \subcaption{Arrow}
\end{minipage}
\centering
\begin{minipage}{\linewidth}
  \centering
  \includegraphics[width=\linewidth]{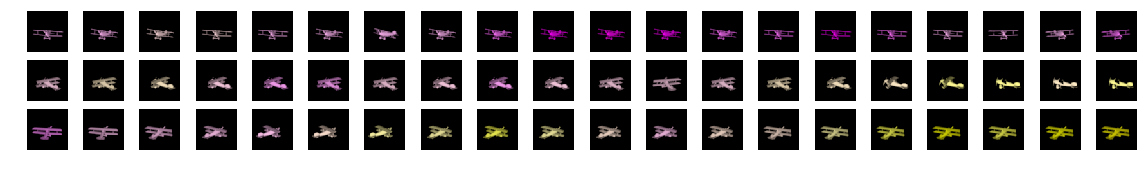}
  \subcaption{Airplane}
\end{minipage}

  \caption{Example paths of consecutive observations.}
  \label{fig:path_examples}
\end{figure*}

The ModelNet40 and the COIL-100 datasets consist of different objects rotating with respect to a vertical axis (out-of-plane rotation). For these datasets the group $G = \mathrm{SO}(2)$ describes the underlying transformations that each object undergoes, see Fig.~\ref{fig:dataset_example2}. The different objects can be seen as non-symmetric variability in the data. In this particular case, each object has its own base-point $x_0$ from which data is generated. The metric \metricname{} is then evaluated per object instance for the group $G = \mathrm{SO}(2)$, the value of  \metricname{} is calculated and averaged across all available objects. Fig.~\ref{fig:path_examples} shows some example paths of consecutive observations for the Square, Arrow, and Airplane datasets, as explained in Sect.~\ref{sec:experimental_setup}.

\paragraph{Square}
This dataset consists of a set of images of a black background with a square of $16\times16$ white pixels. The dataset is generated applying vertical and horizontal translations of the white square considering periodic boundaries.

\paragraph{Arrow}
This dataset consists of a set of images depicting a colored arrow at a given orientation. The dataset is generated by applying cyclic shifts of its color and in-plane rotations. The cyclic color shifts were obtained by preselecting a fixed set of $64$ colors from a circular hue axis. The in-plane rotations were obtained by rotating the arrow along an axis perpendicular to the picture plane over $64$ predefined positions.

\paragraph{Airplane}
This dataset consists of renders obtained using Blender v2.7 \citep{blender2020} from a 3D model of an airplane within the ModelNet40 dataset \citep{Wu2014} (this dataset is provided for the convenience of academic research only). We created each image by varying two properties: the airplane's color and its orientation with respect to the render camera. The orientation was changed via rotation with respect to a vertical axis (out-of-plane rotation). The colors of the model were selected from a predefined cyclic set of colors similar to the arrow rotation dataset.

\paragraph{ModelNet40}
This dataset also consists of a dataset of renders obtained using Blender v2.7 \citep{blender2020} from  the 626 training 3D models within the airplane category of the ModelNet40 dataset \citep{Wu2014}. We created each image by varying each airplane's orientation with respect to the render camera, via rotation with respect to a vertical axis (out-of-plane rotation). In this case we used 64 orientations for each object, i.e. $|\mathcal{G}| = 64$, for a total of 626 objects, thus the dataset consists of 40,064 images.    

\paragraph{COIL-100}
This dataset \citep{nene1996columbia} consists of images from 100 objects placed on a turntable against a black background. For each object, 72 views of the rotated object are provided. The original images have a resolution of $128\times 128$ and were re-scaled to $64\times64$ to match our other datasets. In this case for each object $|\mathcal{G}| = 72$, thus the total dataset consists of 7200 images. This dataset is intended for non-commercial research purposes only. This dataset was obtained using \citet{TFDS}.

\section{Experimental Settings and Hyperparameters}
\label{app:exp_settings}
\subsection{Architectures}
Table~\ref{tab:encoder-decoder} shows the encoder and decoder architectures used for almost all methods and datasets. The encoder's last layer depends on the method. For VAE, cc-VAE, FactorVAE, DIP-I, DIP-II, two dense layers with 4 units each were used. For \methodname{} and $\Delta$VAE two dense layers with 4 and 2 units each were used. For Quessard a single dense layer with 4 units was used. The only model that was not trained with this architectures was \methodname{}/0 method for the ModelNet40 dataset the reason for this choice was that during training the loss was getting NaN values, in this case the architecture used was that of Table~\ref{tab:encoder-decoder-dense}. 

\begin{table*}[ht!]
    \caption{Encoder and decoder architectures used in most methods. }
    \vskip 0.15in
    \begin{center}
\begin{small}
\begin{sc}
    \label{tab:encoder-decoder}
        \begin{tabular}{l l }
            \toprule
            \multicolumn{2}{c|}{Encoder}  \\
            \midrule
            Input& size (64,64, number channels)  \\
            Conv& filters 32, kernel 4,  stride 2, ReLU  \\
            Conv& filters 32, kernel 4,  stride 2, ReLU\\
            Conv& filters 64, kernel 4,  stride 2, ReLU\\
            Conv& filters 64, kernel 4,  stride 2, ReLU\\
            Dense& units 256, ReLU                           \\
            Dense(x2)& units depend on method                                   \\
            \midrule
            \multicolumn{2}{c}{Decoder}\\
            \midrule
              Input & size (number of latent dimensions)\\
            Dense& units 256, ReLU\\
              Dense& units 4*4*64, ReLU\\
              Reshape & (4,4,64)\\
              ConvT& filters 64, kernel 4, stride 2, ReLU\\
              ConvT& filters 32, kernel 4, stride 2, ReLU\\
               ConvT& filters 32, kernel 4, stride 2, ReLU\\
                 ConvT& filters (number channels), kernel 4, stride 2, Sigmoid\\
            
            \bottomrule
            \end{tabular}
\end{sc}
\end{small}
\end{center}
    \vskip -0.1in
\end{table*}

\begin{table*}[ht!]
    \caption{Encoder and decoder architecture used to train \methodname{}/0 for ModelNet40 dataset. }
    \vskip 0.15in
    \begin{center}
\begin{small}
\begin{sc}
    \label{tab:encoder-decoder-dense}
        \begin{tabular}{l l }
            \toprule
            \multicolumn{2}{c|}{Encoder}  \\
            \midrule
            Input& size (64, 64, number channels)  \\
            Dense& units 512, ReLU, batch normalization \\
            Dense& units 256, ReLU, batch normalization \\
            Dense(x2)& units depend on method                                   \\
            \midrule
            \multicolumn{2}{c}{Decoder}\\
            \midrule
              Input & size (number of latent dimensions)\\
              Dense& units 256, ReLU, batch normalization \\
            Dense& units 512, ReLU, batch normalization \\
            Dense& units 64*64*Number of channels, Sigmoid\\
            Reshape& (64, 64, Number of channels)\\

            \bottomrule
            \end{tabular}
\end{sc}
\end{small}
\end{center}
    \vskip -0.1in
\end{table*}

\subsection{Hyperparameters}
Table~\ref{tab:parameters-models} shows the hyperparameters used to train each model for all datasets. Table~\ref{tab:parameters-models-coil} shows the hyperparameters used to train the \methodname{} models for each dataset. In the latter case, the number of epochs for the \methodname{} model were increased. The range of values used for the scale parameter $t$ were increased for ModelNet40 and COIL-100 datasets since it was noticed that this provided better results in terms of data reconstruction and disentanglement. For the Arrow dataset, a value of $\gamma=1$ was producing unstable results. However, the values 10, 100, 1000 or even 10000 were producing good results without significant changes among them. Therefore the value 100 was used for the datasets with the same structure (Square, Arrow and Airplane). For the ModelNet40 and COIL-100 the experiments showed that this hyperparameter for values as high as 10000 could affect the reconstructions, thus a lower value $\gamma=1$ was chosen.

The training of the weakly-supervised models AdaGVAE and AdaMLVAE was done with a data generator that organized the available training data into pairs. The only condition introduced in \citet{Locatello2020} to train these models was to provide paired data with few factors changing among them. For our datasets, two factors change.

\begin{table*}[ht!]
\caption{Model hyperparameters for all datasets}
\label{tab:parameters-models}
\vskip 0.15in
\begin{center}
\begin{small}
\begin{sc}
\begin{tabular}{lc}
\toprule
Model & Parameters\\
\midrule
VAE    & training steps 30000 \\
$\beta$-VAE    & $\beta = 5$, training steps 30000 \\
cc-VAE    & $\beta=5$,$\gamma = 1000$, $c_{max} = 15$, iteration threshold 3500, training steps 30000\\
Factor &     $\gamma = 1$, epochs 30000 \\
DIP-I & $\lambda_{od} = 1$, $\lambda_{d} = 10$, training steps 30000\\
DIP-II    & $\lambda_{od} = 1$, $\lambda_{d} = 1$, training steps 30000  \\
AdaGVAE & $\beta=1$, epochs 500\\
AdaMLVAE & $\beta=1$, epochs 500\\
Quessard & $\lambda = 0.01$, trajectories 3000 \\
\bottomrule
\end{tabular}
\end{sc}
\end{small}
\end{center}
\vskip -0.1in
\end{table*}

\begin{table*}[ht!]
\caption{\methodname{} hyperparameters for all datasets}
\label{tab:parameters-models-coil}
\vskip 0.15in
\begin{center}
\begin{small}
\begin{sc}
\begin{tabular}{lc}
\toprule
 Datasets& Parameters\\
\midrule

Square, Arrow, Airplane & $t\in [10^{-10},     10^{-9}]$, $\gamma=100.0$, epochs 1500 \\
 ModelNet40& $t\in [10^{-10},     10^{-5}]$, $\gamma=1.0$, epochs 1500 \\
 COIL-100& $t\in [10^{-10},     10^{-5}]$, $\gamma=1.0$, epochs 6000 \\

\bottomrule
\end{tabular}
\end{sc}
\end{small}
\end{center}
\vskip -0.1in
\end{table*}

\subsection{Hardware \& Running Time}
The hardware used across all experiments was a DGX station with 4 NVIDIA GPUs V100 and 32GB . Only one GPU was used per experiment. The running time for the \methodname{} across all 9 degrees of supervision $L\in\{0, 256, 768, 1024, 1280, 1536, 1792, 2048\}$ and all 10 runs (total $9 \cdot 10$ repetitions) for the datasets were:  Arrow $33\pm 4$ minutes Airplane $29\pm 4$ minutes and Square $28\pm 4$ minutes. The running time for the \methodname{} across 2 degrees of supervision and 10 runs (total $2 \cdot 10$ repetitions) for ModelNet40 was $136\pm10$ minutes and for COIL-100
$90\pm6$ minutes. For the method from \citep{Quessard2020} the training times were approximately 30 minutes across all datasets. The training times for the methods from \texttt{disentanglement\_lib} \citep{locatello2019challenging} were not measured. 

\subsection{Code Licenses}

The \texttt{disentanglement\_lib} \citep{locatello2019challenging} code is registered with an Apache 2.0 License while the code used to reproduce the method by \citet{Quessard2020} is registered with an MIT license.

\section{Qualitative Results}
\label{app:qualitative_results}

\subsection{Data Generation}
\label{app:data_generation}
\begin{figure*}[ht!]
\centering
\begin{minipage}{0.49\linewidth}
  \centering
  \includegraphics[width=\linewidth]{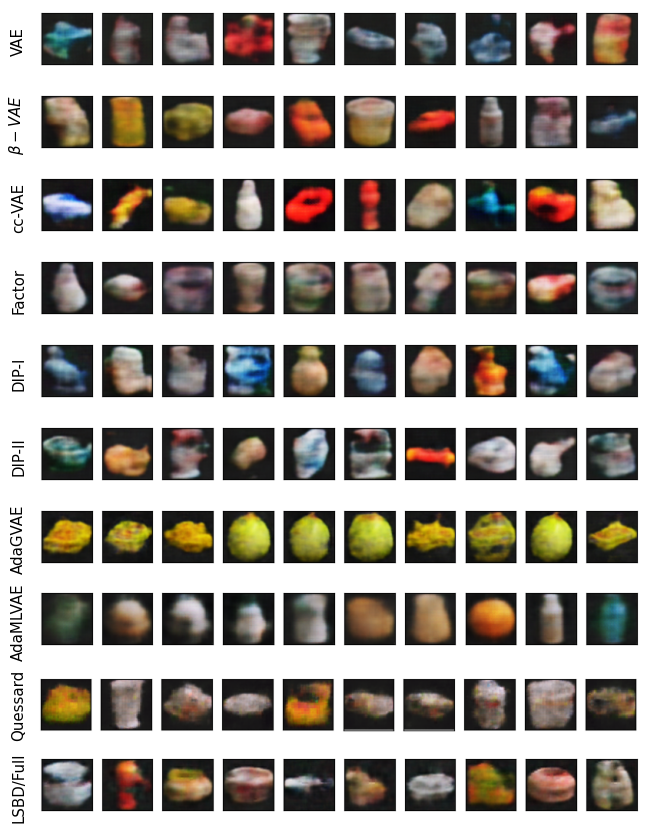}
  \subcaption{COIL 100}
\end{minipage}
\centering
\begin{minipage}{0.49\linewidth}
  \centering
  \includegraphics[width=\linewidth]{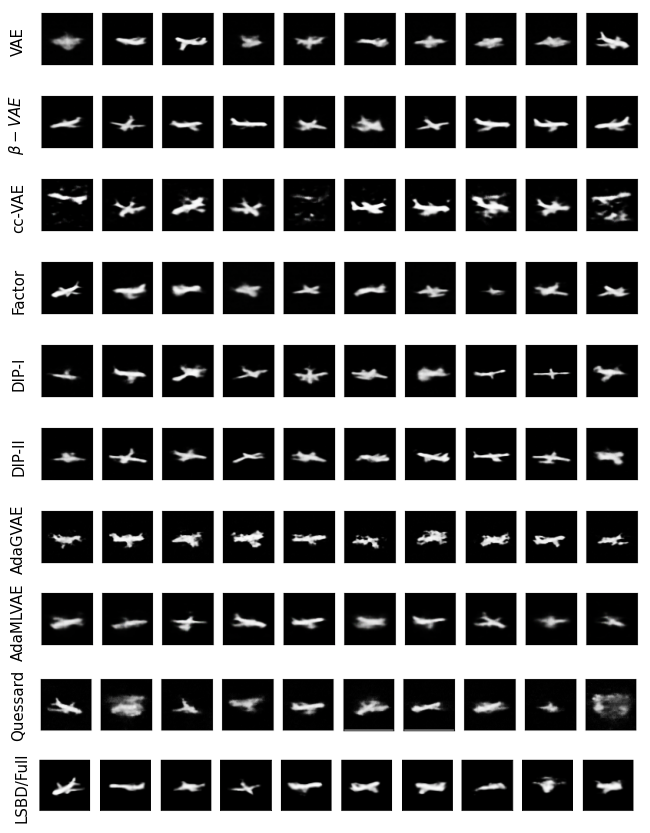}
  \subcaption{ModelNet40 Airplanes}
\end{minipage}
  \caption{Images obtained by decoding latent variables sampled according to the prior over the latent space for different models trained on the COIL-100 and ModelNet40 airplanes datasets.}
  \label{fig:prior_generation}
\end{figure*}

Inspecting data generated by a model can help understand the structure of the learnt latent space in a qualitative way. Fig.~\ref{fig:prior_generation} shows generated data obtained by sampling and decoding ten latent variables for each of the models trained on the COIL-100 and ModelNet40 airplanes datasets. Each latent variable is sampled from the prior over the latent space and decoded to produce an image.

In general, all models but one produce similar results consisting of objects with unclear shape or identity. It is important to highlight the AdaGVAE weakly-supervised model trained on COIL-100 since it appears to have a degenerate decoder producing only yellow objects. Such behaviour occurs for all ten trained instances of the AdaGVAE model. 

Even though the randomly generated images seem to have no clear identity or shape for COIL-100, \methodname{} allows to better determine the identity of such sampled models, by showing multiple orientations thanks to the structure of its latent space. \methodname{} uses a latent space combining an $S^1$ manifold encouraged to encode information about the $SO(2)$ rotations and an Euclidean latent space encouraged to represent the information about the object's identity.

\begin{figure*}[ht!]
\centering
\begin{minipage}{0.43\linewidth}
  \includegraphics[width=\linewidth]{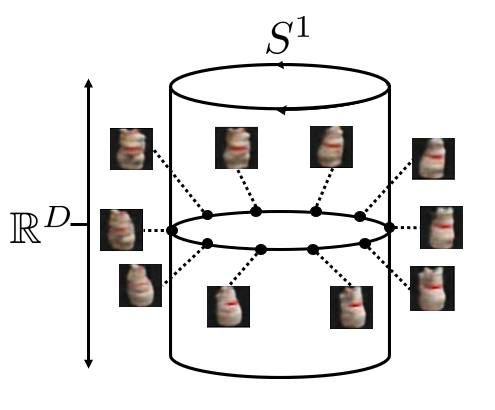}
  \subcaption[]{Latent space structure}
\end{minipage}
  \centering
 \begin{minipage}{0.40\linewidth}
  \includegraphics[width=\linewidth]{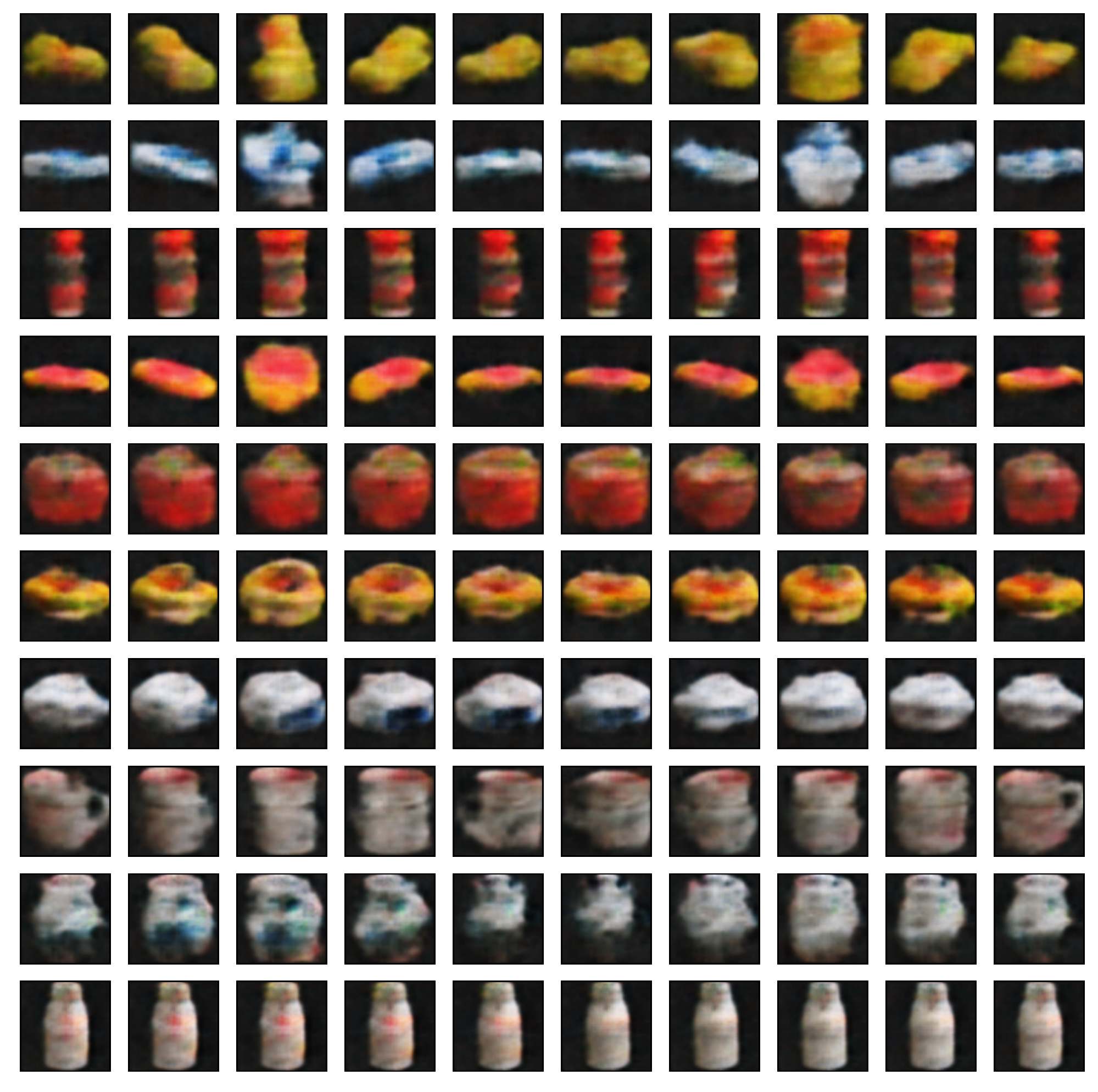}
  
  \subcaption[]{Generated data}
  \end{minipage}

  \caption{Image generation by traversing the circular latent variable for a sampled object identity. The high dimensional Euclidean space is depicted as a single dimension in a hyper-cylinder. (a) The latent variable corresponding to the identity is sampled from the prior over the Euclidean latent space and combined with regularly spaced latent variables on $S^1$. (2) Each row presents the decoded images for a fixed Euclidean latent variable while each column shows the images for a fixed latent variable on $S^1$. The images are obtained from decoding the latent variables with \methodname{}/full trained on the COIL-100 dataset.}
  \label{fig:cond_generation}
\end{figure*}

By first sampling a latent variable from the Euclidean latent space and combining it with a set of regularly spaced latent variables along $S^1$ we can observe some consistency in object identity, see Fig.~\ref{fig:cond_generation}. Such data generation cannot be directly obtained from traditional disentanglement methods since there is no clear direction representing either the object identities or the orientations. 


\subsection{Object Interpolation}
\label{app:latent_traversals}

Next, we will show how the latent space is structured among the latent variables representing the objects' identities for models trained with COIL-100. We show the generated data obtained from decoding linearly interpolated latent variables between different objects to show the transitions between objects and orientations.

\begin{figure*}[ht!]
\centering
\begin{minipage}{0.40\linewidth}
  \includegraphics[width=\linewidth]{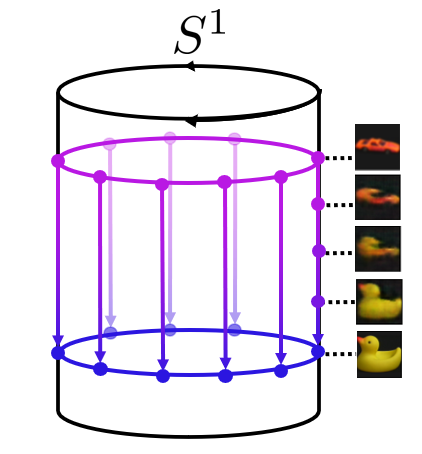}
  \subcaption[]{Interpolation across $Z = S^1\times \mathbb{R}^D$}
\end{minipage}
  \centering
 \begin{minipage}{0.40\linewidth}
  \includegraphics[width=\linewidth]{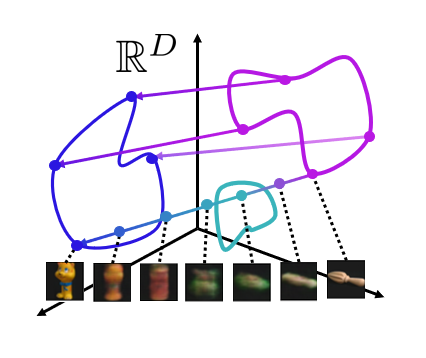}
  
  \subcaption[]{Interpolation across $Z = \mathbb{R}^D$}
  \label{fig:linear_interpolation}
  \end{minipage}

  \caption{Diagrams illustrating the interpolation between the latent variables associated to two objects. (a) Interpolation across a hyper-cylinder within $Z = S^1\times \mathbb{R}^D$ used by \methodname{}. (b) Interpolation across $Z = \mathbb{R}^D$ of traditional disentanglement models. In the traditional disentanglement models the linear interpolation can show the crossing of the latent codes associated to unexpected objects.}
  \label{fig:traversal_manifolds}
\end{figure*}

For \methodname{} the interpolation is simple; first the latent variables associated to the identity of the start and end objects are estimated by averaging the Euclidean latent variables of all images per object. Second, the linear interpolation between the object identity latent variables of the start and end object is calculated to generate a path through the object identity space. Finally, the estimated identity variables in the path are combined with regularly spaced variables in the orientation space $S^1$ and decoded. See Fig.~\ref{fig:linear_interpolation} (a). 

In the case of the traditional disentanglement methods we cannot produce a latent variable representing an object's identity, so there is no clear traversal between objects. In this case, a linear interpolation between an image from the start object to and end object is calculated and the latent variables are decoded, see Fig.~\ref{fig:linear_interpolation}. Notice that we cannot easily produce an image of an object with an arbitrary orientation since we do not know the shape of the loop in the latent space representing an object. 

\begin{figure*}[ht!]
\centering
\begin{minipage}{0.39\linewidth}
  \centering
  \includegraphics[width=\linewidth]{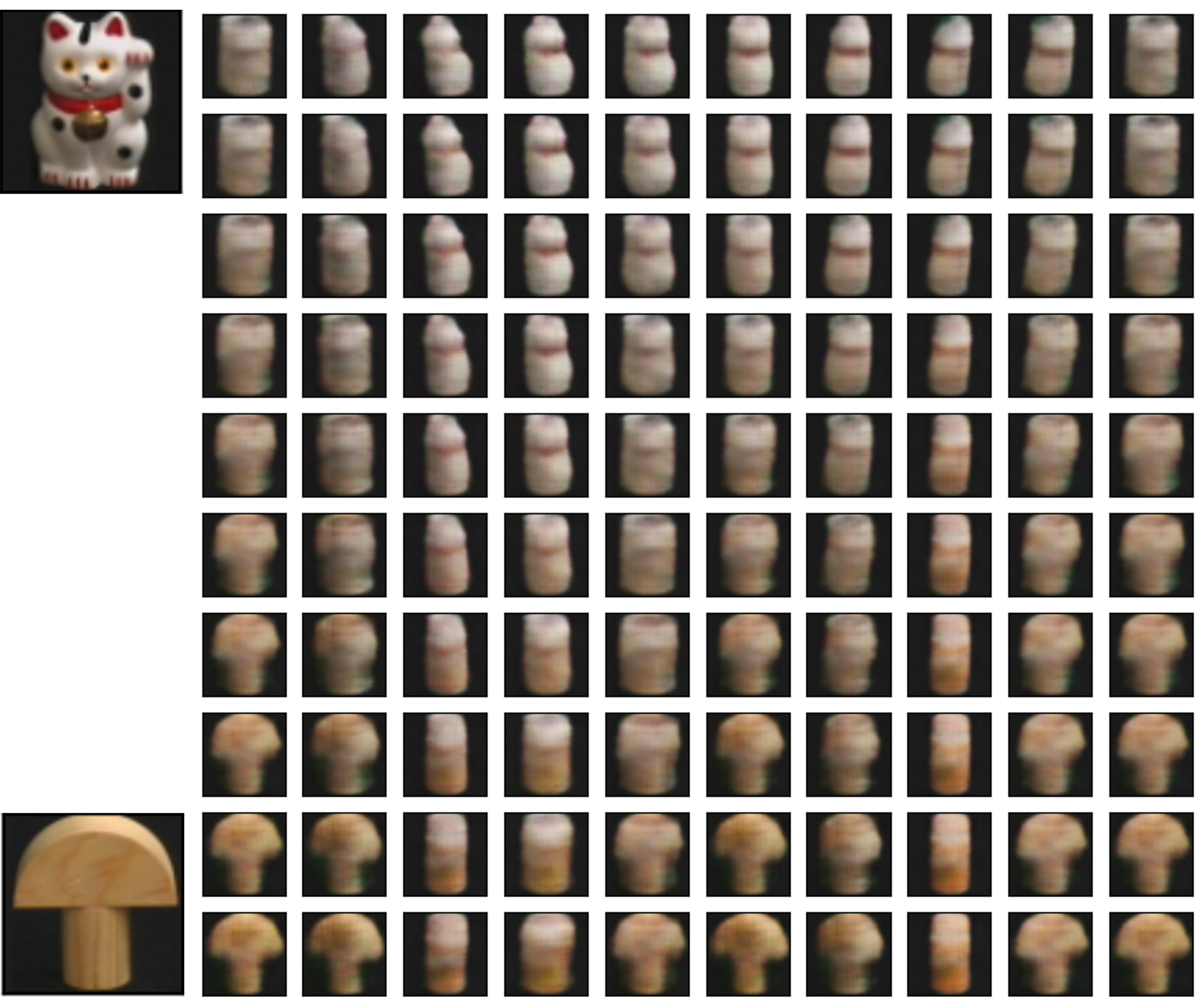}
\end{minipage}
\centering
\begin{minipage}{0.33\linewidth}
  \centering
  \includegraphics[width=\linewidth]{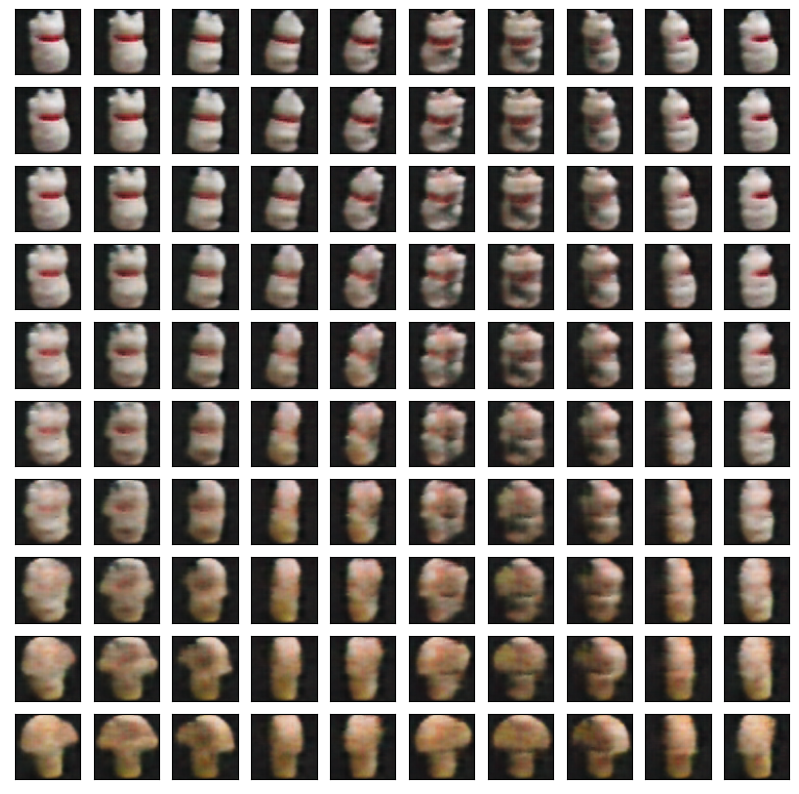}
\end{minipage}

\centering
\begin{minipage}{0.39\linewidth}
  \centering
  \includegraphics[width=\linewidth]{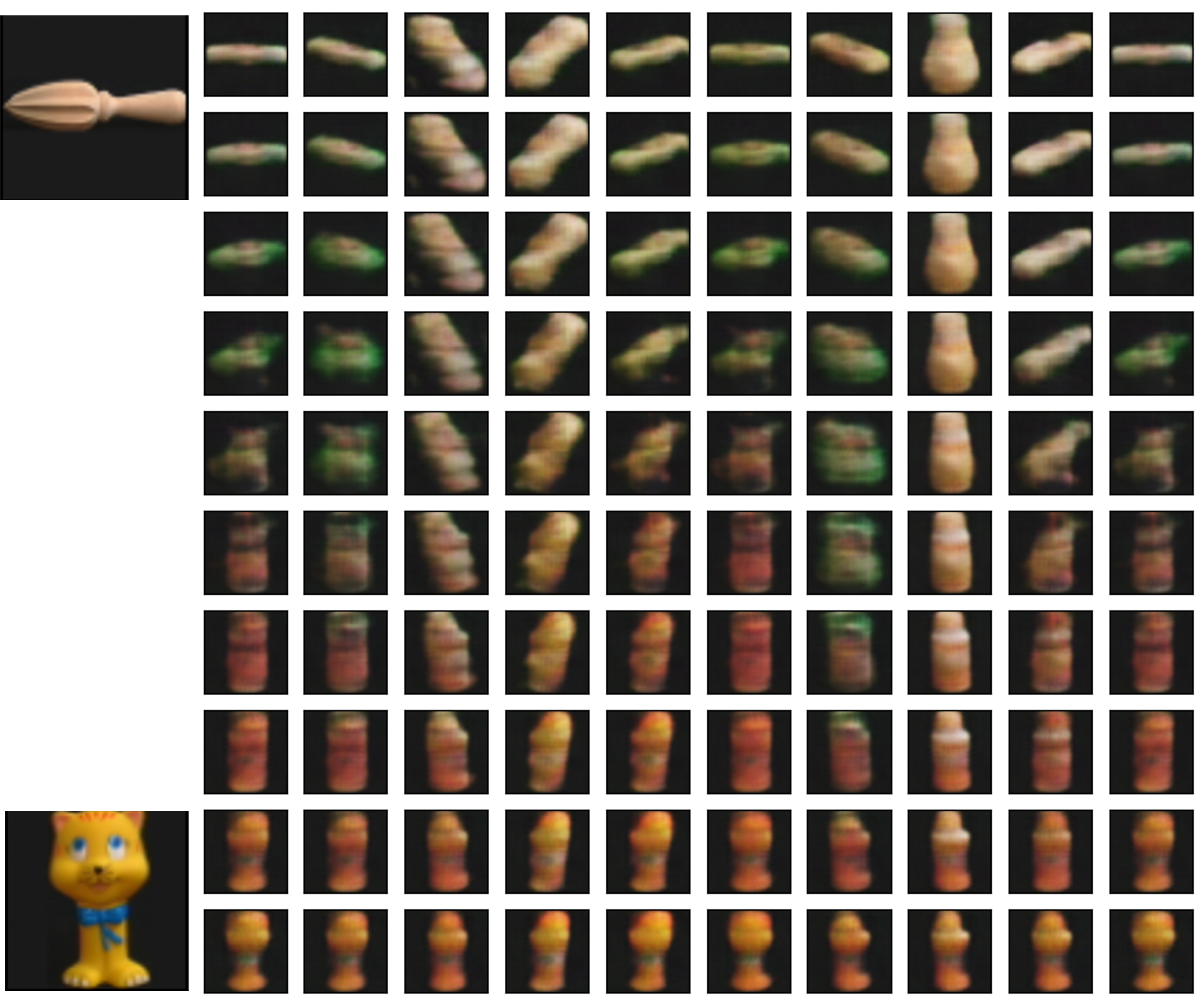}
\end{minipage}
\centering
\begin{minipage}{0.33\linewidth}
  \centering
  \includegraphics[width=\linewidth]{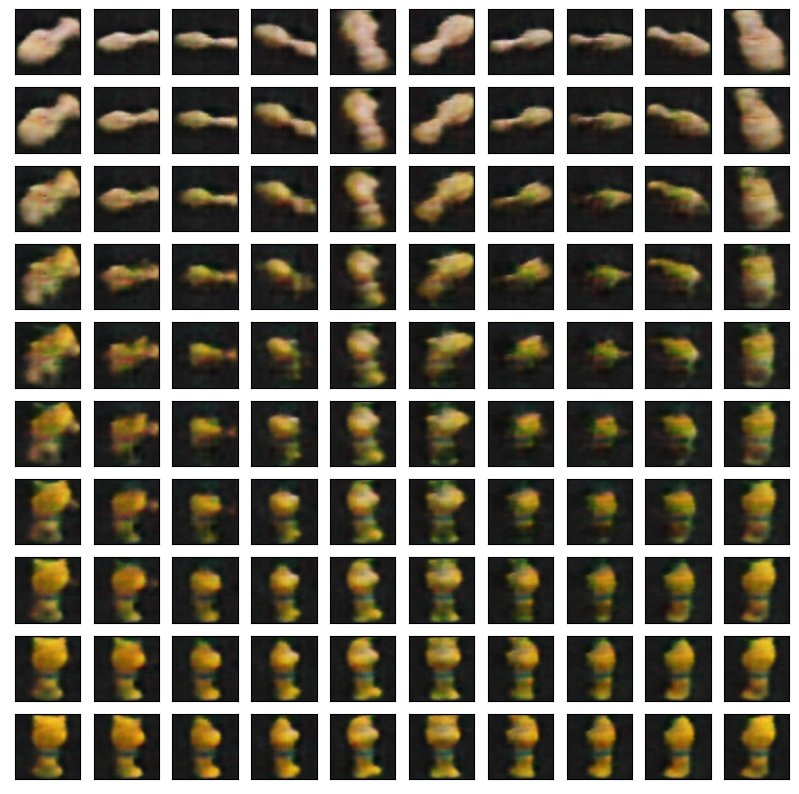}
\end{minipage}

\centering
\begin{minipage}{0.39\linewidth}
  \centering
  \includegraphics[width=\linewidth]{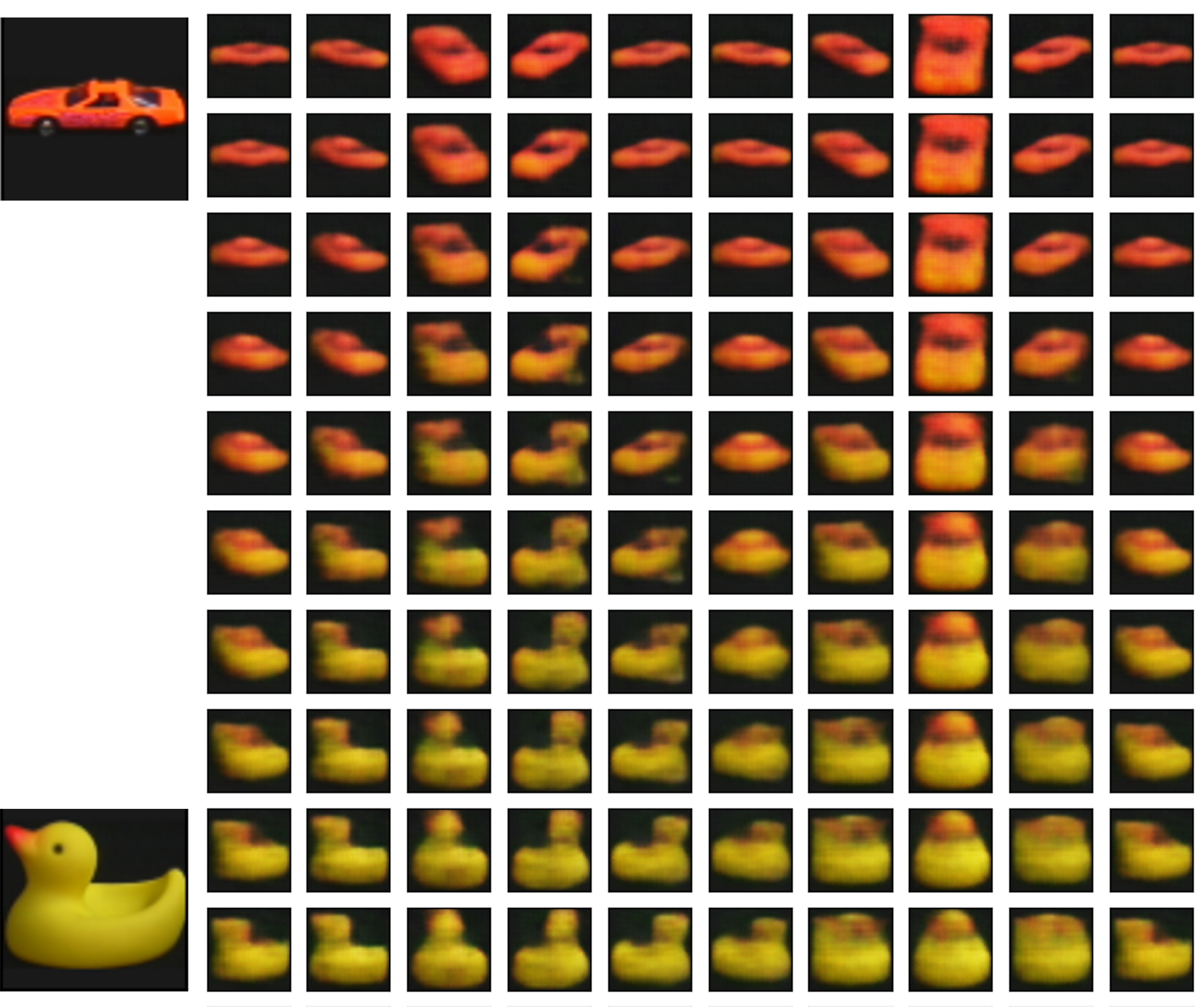}
  \subcaption{  cc-VAE}
\end{minipage}
\centering
\begin{minipage}{0.33\linewidth}
  \centering
  \includegraphics[width=\linewidth]{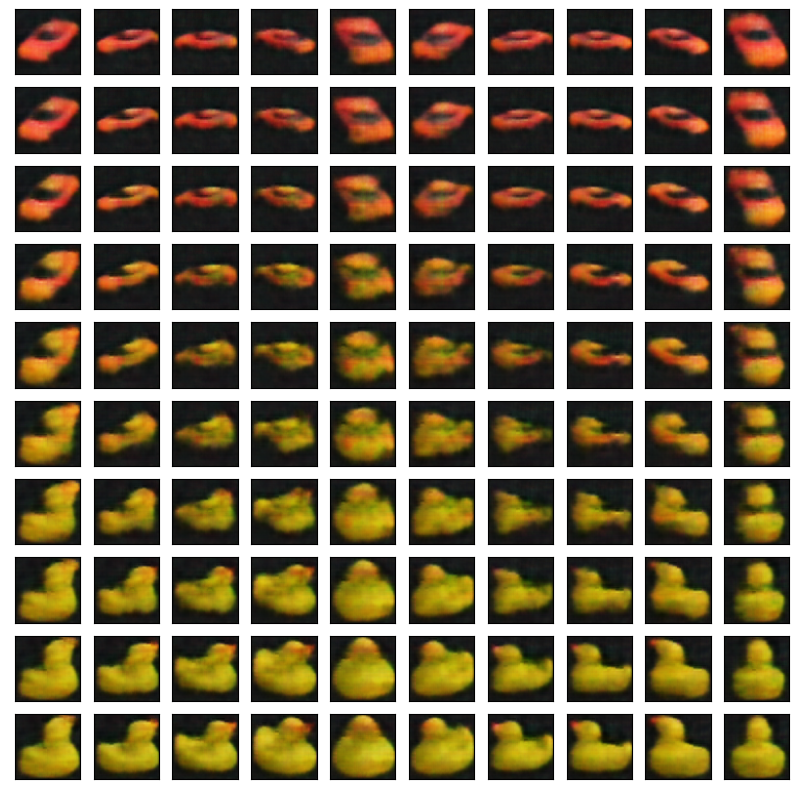}
  \subcaption{\methodname{}}
\end{minipage}

  \caption{Images produced from the decoding of interpolated latent variables using cc-VAE and \methodname{} trained with COIL-100. Three interpolations between two objects are shown. Each column represents the transitions between objects while each row shows images that should correspond to different orientations. }
  \label{fig:obj_interpolation}
\end{figure*}

Fig.~\ref{fig:obj_interpolation} shows the generated images obtained by interpolating between two objects. We only show cc-VAE representing traditional models since that method attained the lowest \metricname{}. A particularly interesting interpolation is between the wooden object and the orange cat figure. The interpolation of cc-VAE shows how a green object is also crossed in between while \methodname{} shows a consistent transition between the objects a visual explanation of this observation is presented in Fig.~\ref{fig:linear_interpolation} (b). 

\newpage

\section{Full results}
\label{app:full_results}
The full results for all experiments on all datasets are given in Tables \ref{tab:results_square}, \ref{tab:results_arrow}, \ref{tab:results_airplane}, \ref{tab:results_modelnet40}, and \ref{tab:results_coil100}. We report the mean and standard deviation over 10 runs for each experiment.

\subsection{Limited Supervision Suffices to Learn LSBD Representations}

\begin{figure*}[!t]
\centering
\begin{minipage}{0.32\textwidth}
  \centering
\includegraphics[width=\textwidth]{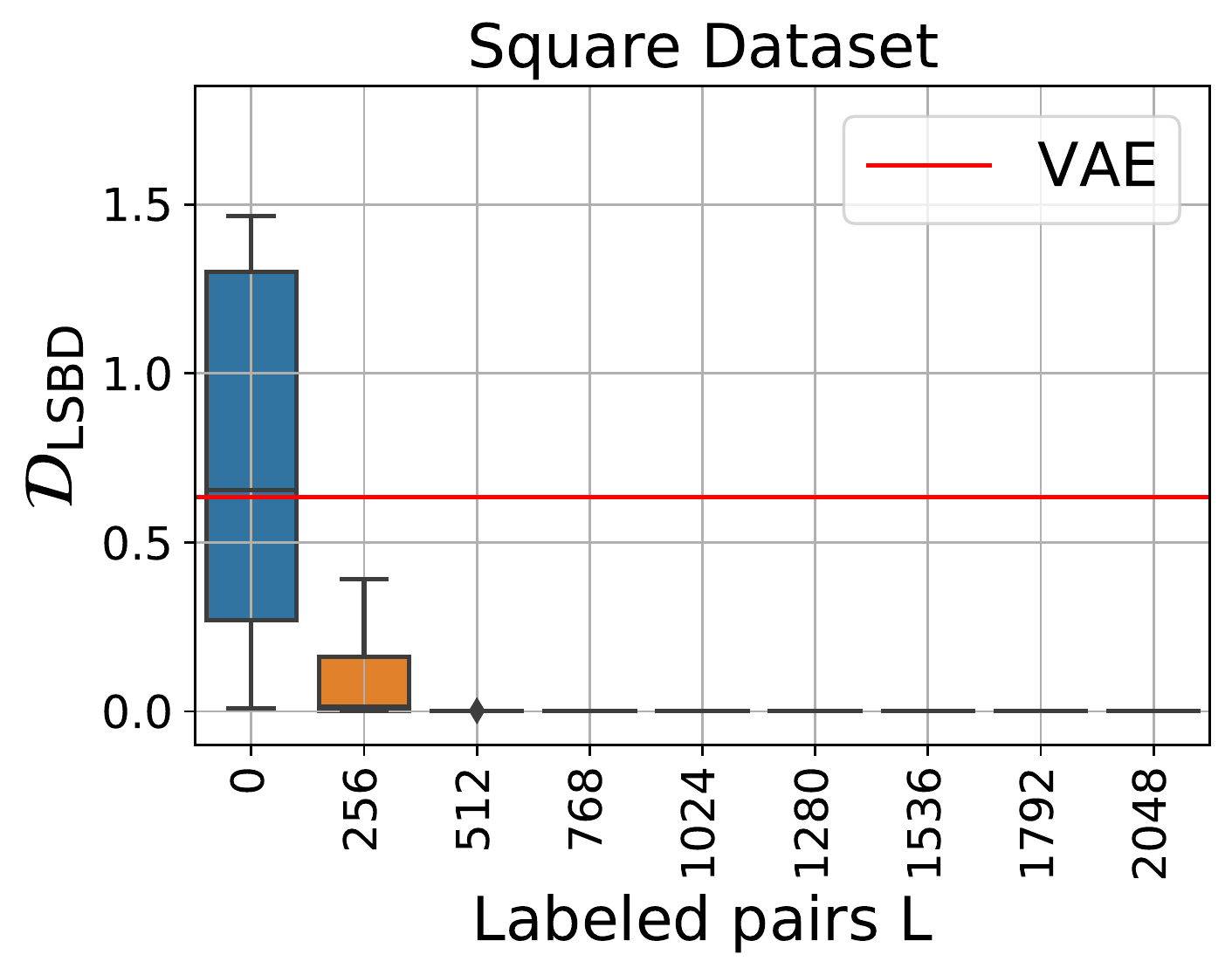}
\end{minipage}
\begin{minipage}{0.32\textwidth}
  \centering
\includegraphics[width=\textwidth]{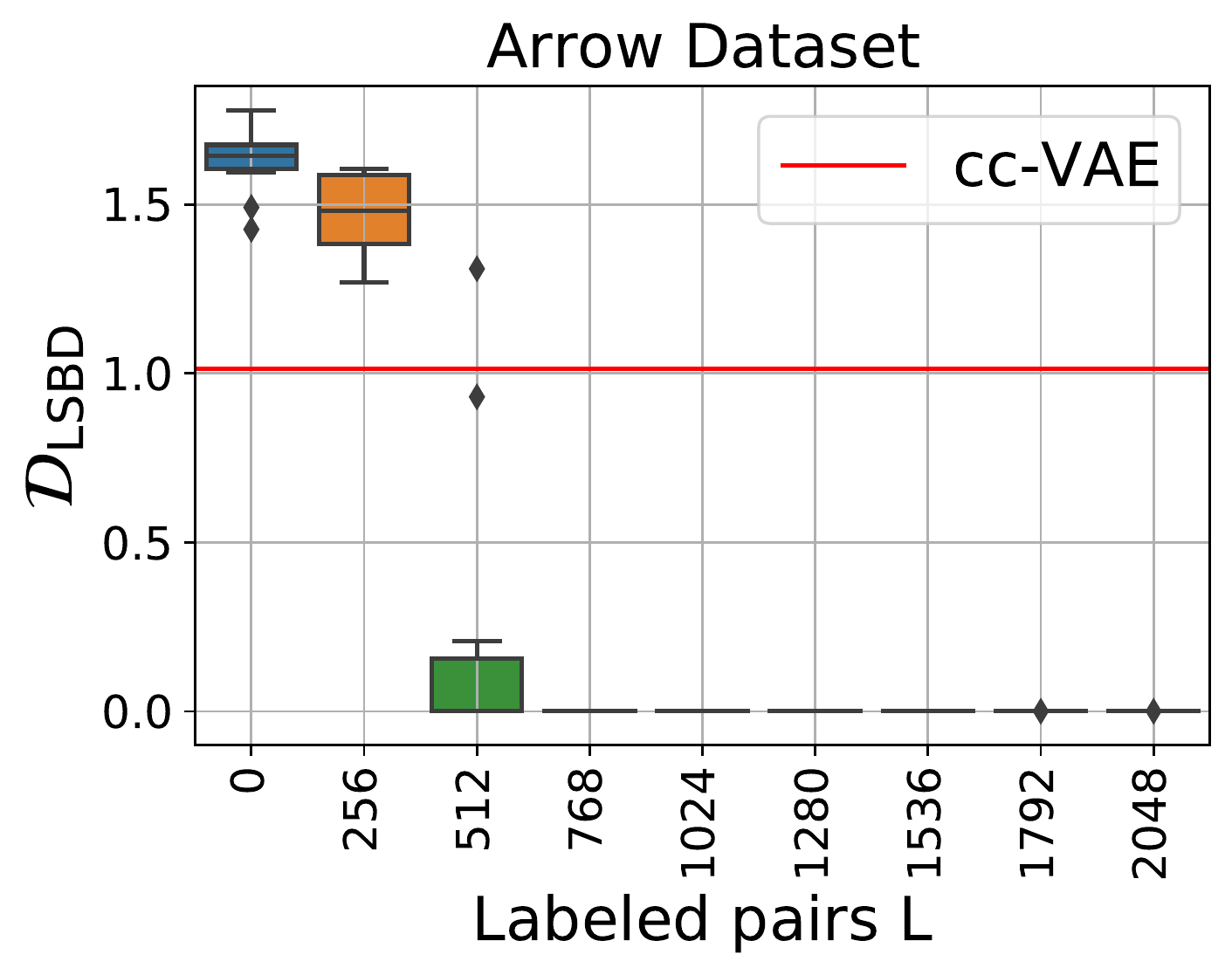}
\end{minipage}
\begin{minipage}{0.32\textwidth}
  \centering
\includegraphics[width=\textwidth]{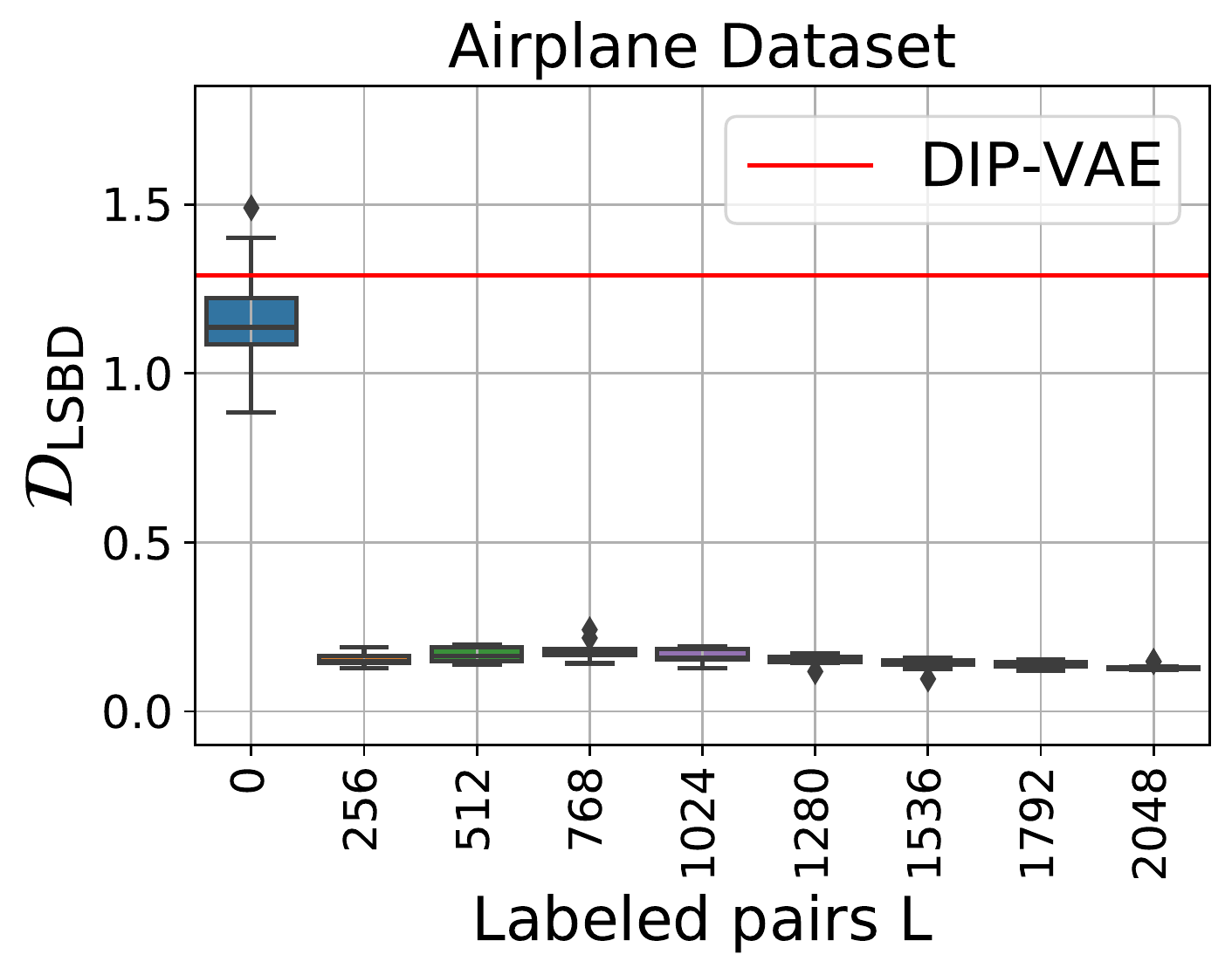}
\end{minipage}
  \caption{Box plots for \metricname{} scores over 10 training repetitions for different numbers of labeled pairs $L$, for all datasets. The red line indicates the best-performing traditional disentanglement method.}
  \label{fig:results_number_of_labels}
\end{figure*}

The results obtained from Tables \ref{tab:results_square}, \ref{tab:results_arrow}, \ref{tab:results_airplane} show that we do not need transformation-labels for all data points, only a subset of labeled pairs is sufficient to learn LSBD representations. To further highlight this, Fig.~\ref{fig:results_number_of_labels} shows \metricname{} scores for \methodname{} trained on the Square, Arrow, and Airplane datasets respectively, for various values for the number of labeled pairs $L$. For each $L$ and each dataset, we trained 10 models so we can report box plots of the \metricname{} scores.

For low values of $L$ we see worse scores and high variability. But for slightly higher $L$, scores are consistently good, starting already at $L=512$ for the Square, $L=768$ for the Arrow, and $L=256$ for the Airplane. This corresponds to respectively 25\%, 37.5\%, and 12.5\% of the data being involved in a labeled pair. Moreover, we see that with just a little supervision we outperform the best traditional method on \metricname{}. Overall, these results suggest that with some expert knowledge (about the underlying group and a suitable representation) and limited annotation of transformations, LSBD can be achieved.

\subsection{Quessard Arrow}
\begin{figure*}[ht!]
\centering
\begin{minipage}{0.3\linewidth}
  \centering
  \includegraphics[width=\linewidth]{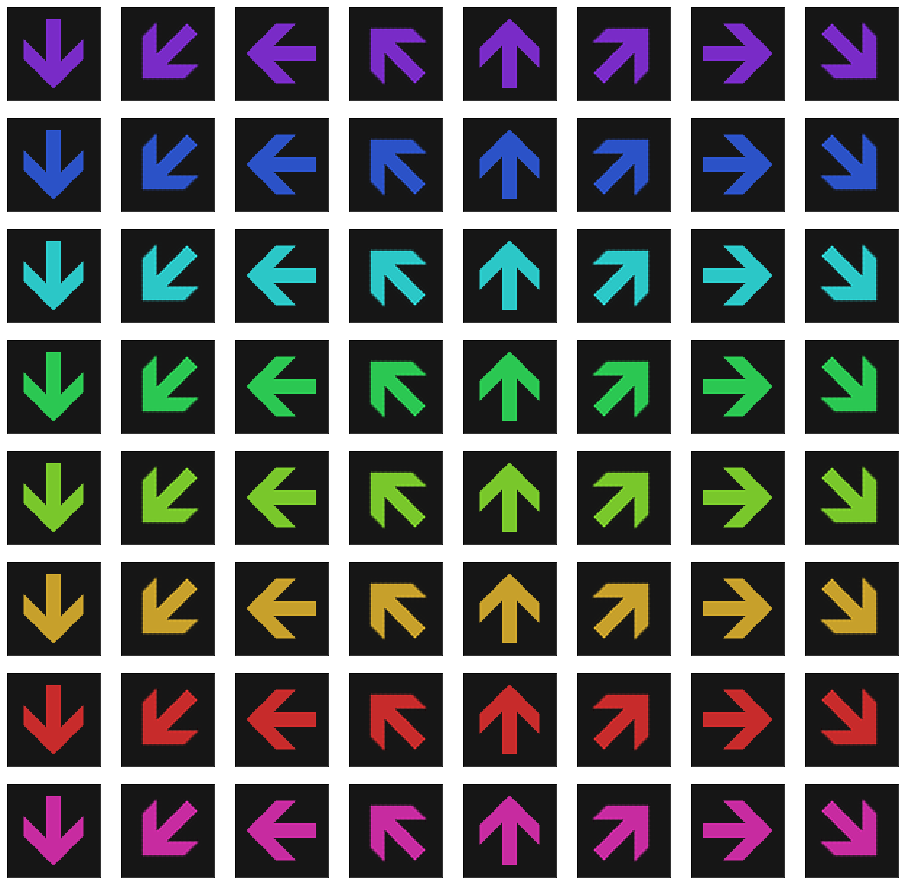}
  \subcaption{Input}
\end{minipage}
\centering
\begin{minipage}{0.3\linewidth}
  \centering
  \includegraphics[width=\linewidth]{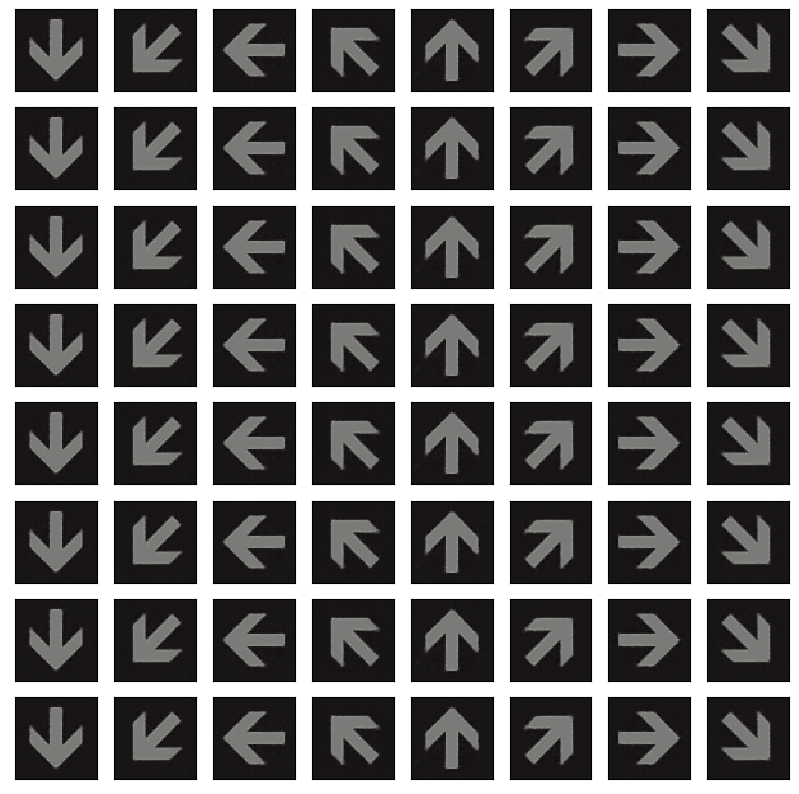}
  \subcaption{Reconstructions}
\end{minipage}
  \caption{Results from \citet{Quessard2020}'s method on the Arrow dataset}
  \label{fig:gray_arrow}
\end{figure*}

In the main text we mentioned that we did not reproduce good results with \citet{Quessard2020}'s method on the Arrow and Square dataset. We highlight a particular case for the Arrow dataset, where the method clearly learns the rotations of the arrow but fails to learn color. Fig.~\ref{fig:gray_arrow} shows reconstructed Arrow images. Since color isn't learned well, this example doesn't get a good \metricname{} score, even though rotation is properly linearly disentangled.

\begin{table*}[hb!]
\caption{Scores for the Square dataset.}
\label{tab:results_square}
\vskip 0.1in
\begin{center}
\begin{small}
\begin{sc}
\begin{tabular}{lccccccc}
\toprule
Model & Beta $\uparrow$ & Factor $\uparrow$ & SAP $\uparrow$ & DCI $\uparrow$ & MIG $\uparrow$ & MOD $\uparrow$ & \metricname{} $\downarrow$\\
\midrule
VAE&\multirow{2}{*}{$.945$\tiny{$\pm.061$}}&\multirow{2}{*}{$.835$\tiny{$\pm.140$}}&\multirow{2}{*}{$.019$\tiny{$\pm.004$}}&\multirow{2}{*}{$.009$\tiny{$\pm.005$}}&\multirow{2}{*}{$.013$\tiny{$\pm.004$}}&\multirow{2}{*}{$.579$\tiny{$\pm.202$}}&\multirow{2}{*}{$.634$\tiny{$\pm.440$}}\\
&&&&&&&\\
$\beta$-VAE&\multirow{2}{*}{$.980$\tiny{$\pm.033$}}&\multirow{2}{*}{$.913$\tiny{$\pm.095$}}&\multirow{2}{*}{$.021$\tiny{$\pm.006$}}&\multirow{2}{*}{$.017$\tiny{$\pm.011$}}&\multirow{2}{*}{$.021$\tiny{$\pm.014$}}&\multirow{2}{*}{$.642$\tiny{$\pm.147$}}&\multirow{2}{*}{$.732$\tiny{$\pm.488$}}\\
&&&&&&&\\
cc-VAE&\multirow{2}{*}{$.508$\tiny{$\pm.023$}}&\multirow{2}{*}{$.000$\tiny{$\pm.000$}}&\multirow{2}{*}{$.003$\tiny{$\pm.002$}}&\multirow{2}{*}{$.007$\tiny{$\pm.002$}}&\multirow{2}{*}{$.014$\tiny{$\pm.004$}}&\multirow{2}{*}{$.222$\tiny{$\pm.110$}}&\multirow{2}{*}{$1.905$\tiny{$\pm.023$}}\\
&&&&&&&\\
Factor&\multirow{2}{*}{$.974$\tiny{$\pm.048$}}&\multirow{2}{*}{$.910$\tiny{$\pm.104$}}&\multirow{2}{*}{$.020$\tiny{$\pm.003$}}&\multirow{2}{*}{$.019$\tiny{$\pm.017$}}&\multirow{2}{*}{$.017$\tiny{$\pm.010$}}&\multirow{2}{*}{$.712$\tiny{$\pm.183$}}&\multirow{2}{*}{$.667$\tiny{$\pm.428$}}\\
&&&&&&&\\
DIP-I&\multirow{2}{*}{$.972$\tiny{$\pm.042$}}&\multirow{2}{*}{$.861$\tiny{$\pm.097$}}&\multirow{2}{*}{$.020$\tiny{$\pm.005$}}&\multirow{2}{*}{$.010$\tiny{$\pm.002$}}&\multirow{2}{*}{$.011$\tiny{$\pm.002$}}&\multirow{2}{*}{$.618$\tiny{$\pm.117$}}&\multirow{2}{*}{$1.109$\tiny{$\pm.312$}}\\
&&&&&&&\\
DIP-II&\multirow{2}{*}{$.930$\tiny{$\pm.119$}}&\multirow{2}{*}{$.848$\tiny{$\pm.137$}}&\multirow{2}{*}{$.018$\tiny{$\pm.004$}}&\multirow{2}{*}{$.010$\tiny{$\pm.004$}}&\multirow{2}{*}{$.015$\tiny{$\pm.007$}}&\multirow{2}{*}{$.607$\tiny{$\pm.207$}}&\multirow{2}{*}{$.907$\tiny{$\pm.559$}}\\
&&&&&&&\\
AdaGVAE&\multirow{2}{*}{$.841$\tiny{$\pm.230$}}&\multirow{2}{*}{$.707$\tiny{$\pm.386$}}&\multirow{2}{*}{$.009$\tiny{$\pm.009$}}&\multirow{2}{*}{$.024$\tiny{$\pm.015$}}&\multirow{2}{*}{$.012$\tiny{$\pm.005$}}&\multirow{2}{*}{$.473$\tiny{$\pm.185$}}&\multirow{2}{*}{$.666$\tiny{$\pm.378$}}\\
&&&&&&&\\
AdaMLVAE&\multirow{2}{*}{$.737$\tiny{$\pm.208$}}&\multirow{2}{*}{$.465$\tiny{$\pm.403$}}&\multirow{2}{*}{$.008$\tiny{$\pm.008$}}&\multirow{2}{*}{$.016$\tiny{$\pm.006$}}&\multirow{2}{*}{$.013$\tiny{$\pm.007$}}&\multirow{2}{*}{$.338$\tiny{$\pm.128$}}&\multirow{2}{*}{$1.063$\tiny{$\pm.387$}}\\
&&&&&&&\\
Quessard&\multirow{2}{*}{$.504$\tiny{$\pm.021$}}&\multirow{2}{*}{$.000$\tiny{$\pm.000$}}&\multirow{2}{*}{$.004$\tiny{$\pm.003$}}&\multirow{2}{*}{$.007$\tiny{$\pm.004$}}&\multirow{2}{*}{$.018$\tiny{$\pm.008$}}&\multirow{2}{*}{$.354$\tiny{$\pm.213$}}&\multirow{2}{*}{$1.686$\tiny{$\pm.294$}}\\
&&&&&&&\\

\methodname{}&\multirow{2}{*}{$.970$\tiny{$\pm.079$}}&\multirow{2}{*}{$.913$\tiny{$\pm.121$}}&\multirow{2}{*}{$.018$\tiny{$\pm.003$}}&\multirow{2}{*}{$.052$\tiny{$\pm.052$}}&\multirow{2}{*}{$.018$\tiny{$\pm.004$}}&\multirow{2}{*}{$.884$\tiny{$\pm.183$}}&\multirow{2}{*}{$.749$\tiny{$\pm.554$}}\\
/0&&&&&&&\\
\methodname{}&\multirow{2}{*}{$1.000$\tiny{$\pm.000$}}&\multirow{2}{*}{$1.000$\tiny{$\pm.001$}}&\multirow{2}{*}{$.021$\tiny{$\pm.004$}}&\multirow{2}{*}{$.267$\tiny{$\pm.152$}}&\multirow{2}{*}{$.027$\tiny{$\pm.007$}}&\multirow{2}{*}{$.986$\tiny{$\pm.023$}}&\multirow{2}{*}{$.104$\tiny{$\pm.147$}}\\
/256&&&&&&&\\
\methodname{}&\multirow{2}{*}{$1.000$\tiny{$\pm.000$}}&\multirow{2}{*}{$1.000$\tiny{$\pm.000$}}&\multirow{2}{*}{$.021$\tiny{$\pm.006$}}&\multirow{2}{*}{$.393$\tiny{$\pm.022$}}&\multirow{2}{*}{$.025$\tiny{$\pm.005$}}&\multirow{2}{*}{$.999$\tiny{$\pm.000$}}&\multirow{2}{*}{$.000$\tiny{$\pm.000$}}\\
/512&&&&&&&\\
\methodname{}&\multirow{2}{*}{$1.000$\tiny{$\pm.000$}}&\multirow{2}{*}{$1.000$\tiny{$\pm.000$}}&\multirow{2}{*}{$.019$\tiny{$\pm.004$}}&\multirow{2}{*}{$.387$\tiny{$\pm.014$}}&\multirow{2}{*}{$.025$\tiny{$\pm.004$}}&\multirow{2}{*}{$.999$\tiny{$\pm.000$}}&\multirow{2}{*}{$.000$\tiny{$\pm.000$}}\\
/768&&&&&&&\\
\methodname{}&\multirow{2}{*}{$1.000$\tiny{$\pm.000$}}&\multirow{2}{*}{$1.000$\tiny{$\pm.000$}}&\multirow{2}{*}{$.022$\tiny{$\pm.005$}}&\multirow{2}{*}{$.398$\tiny{$\pm.020$}}&\multirow{2}{*}{$.024$\tiny{$\pm.003$}}&\multirow{2}{*}{$.999$\tiny{$\pm.000$}}&\multirow{2}{*}{$.000$\tiny{$\pm.000$}}\\
/1024&&&&&&&\\
\methodname{}&\multirow{2}{*}{$1.000$\tiny{$\pm.000$}}&\multirow{2}{*}{$1.000$\tiny{$\pm.000$}}&\multirow{2}{*}{$.023$\tiny{$\pm.003$}}&\multirow{2}{*}{$.389$\tiny{$\pm.016$}}&\multirow{2}{*}{$.023$\tiny{$\pm.003$}}&\multirow{2}{*}{$.999$\tiny{$\pm.000$}}&\multirow{2}{*}{$.000$\tiny{$\pm.000$}}\\
/1280&&&&&&&\\
\methodname{}&\multirow{2}{*}{$1.000$\tiny{$\pm.000$}}&\multirow{2}{*}{$1.000$\tiny{$\pm.000$}}&\multirow{2}{*}{$.022$\tiny{$\pm.004$}}&\multirow{2}{*}{$.398$\tiny{$\pm.013$}}&\multirow{2}{*}{$.027$\tiny{$\pm.002$}}&\multirow{2}{*}{$.999$\tiny{$\pm.000$}}&\multirow{2}{*}{$.000$\tiny{$\pm.000$}}\\
/1536&&&&&&&\\
\methodname{}&\multirow{2}{*}{$1.000$\tiny{$\pm.000$}}&\multirow{2}{*}{$1.000$\tiny{$\pm.000$}}&\multirow{2}{*}{$.020$\tiny{$\pm.004$}}&\multirow{2}{*}{$.397$\tiny{$\pm.016$}}&\multirow{2}{*}{$.027$\tiny{$\pm.005$}}&\multirow{2}{*}{$.999$\tiny{$\pm.000$}}&\multirow{2}{*}{$.000$\tiny{$\pm.000$}}\\
/1792&&&&&&&\\
\methodname{}&\multirow{2}{*}{$1.000$\tiny{$\pm.000$}}&\multirow{2}{*}{$1.000$\tiny{$\pm.000$}}&\multirow{2}{*}{$.021$\tiny{$\pm.006$}}&\multirow{2}{*}{$.380$\tiny{$\pm.027$}}&\multirow{2}{*}{$.027$\tiny{$\pm.005$}}&\multirow{2}{*}{$.999$\tiny{$\pm.000$}}&\multirow{2}{*}{$.000$\tiny{$\pm.000$}}\\
/full&&&&&&&\\
\methodname{}&&&&&&&\multirow{2}{*}{$.005$\tiny{$\pm.002$}}\\
 /paths&&&&&&&\\
\bottomrule
\end{tabular}
\end{sc}
\end{small}
\end{center}
\vskip -0.1in
\end{table*}

\begin{table*}[!t]
\caption{Scores for the Arrow dataset.}
\label{tab:results_arrow}
\vskip 0.1in
\begin{center}
\begin{small}
\begin{sc}
\begin{tabular}{lccccccc}
\toprule
Model & Beta $\uparrow$ & Factor $\uparrow$ & SAP $\uparrow$ & DCI $\uparrow$ & MIG $\uparrow$ & MOD $\uparrow$ & \metricname{} $\downarrow$\\
\midrule
VAE&\multirow{2}{*}{$1.000$\tiny{$\pm.000$}}&\multirow{2}{*}{$.646$\tiny{$\pm.032$}}&\multirow{2}{*}{$.017$\tiny{$\pm.004$}}&\multirow{2}{*}{$.009$\tiny{$\pm.003$}}&\multirow{2}{*}{$.013$\tiny{$\pm.004$}}&\multirow{2}{*}{$.961$\tiny{$\pm.012$}}&\multirow{2}{*}{$1.316$\tiny{$\pm.193$}}\\
&&&&&&&\\
$\beta$-VAE&\multirow{2}{*}{$.999$\tiny{$\pm.002$}}&\multirow{2}{*}{$.588$\tiny{$\pm.045$}}&\multirow{2}{*}{$.018$\tiny{$\pm.004$}}&\multirow{2}{*}{$.008$\tiny{$\pm.002$}}&\multirow{2}{*}{$.015$\tiny{$\pm.005$}}&\multirow{2}{*}{$.898$\tiny{$\pm.032$}}&\multirow{2}{*}{$1.178$\tiny{$\pm.065$}}\\
&&&&&&&\\
cc-VAE&\multirow{2}{*}{$.982$\tiny{$\pm.056$}}&\multirow{2}{*}{$.707$\tiny{$\pm.102$}}&\multirow{2}{*}{$.019$\tiny{$\pm.004$}}&\multirow{2}{*}{$.011$\tiny{$\pm.005$}}&\multirow{2}{*}{$.016$\tiny{$\pm.004$}}&\multirow{2}{*}{$.980$\tiny{$\pm.038$}}&\multirow{2}{*}{$1.013$\tiny{$\pm.096$}}\\
&&&&&&&\\
Factor&\multirow{2}{*}{$1.000$\tiny{$\pm.000$}}&\multirow{2}{*}{$.659$\tiny{$\pm.028$}}&\multirow{2}{*}{$.017$\tiny{$\pm.003$}}&\multirow{2}{*}{$.008$\tiny{$\pm.003$}}&\multirow{2}{*}{$.014$\tiny{$\pm.002$}}&\multirow{2}{*}{$.935$\tiny{$\pm.037$}}&\multirow{2}{*}{$1.526$\tiny{$\pm.125$}}\\
&&&&&&&\\
DIP-I&\multirow{2}{*}{$1.000$\tiny{$\pm.000$}}&\multirow{2}{*}{$.624$\tiny{$\pm.042$}}&\multirow{2}{*}{$.020$\tiny{$\pm.004$}}&\multirow{2}{*}{$.008$\tiny{$\pm.002$}}&\multirow{2}{*}{$.012$\tiny{$\pm.003$}}&\multirow{2}{*}{$.967$\tiny{$\pm.027$}}&\multirow{2}{*}{$1.521$\tiny{$\pm.113$}}\\
&&&&&&&\\
DIP-II&\multirow{2}{*}{$1.000$\tiny{$\pm.000$}}&\multirow{2}{*}{$.644$\tiny{$\pm.064$}}&\multirow{2}{*}{$.020$\tiny{$\pm.004$}}&\multirow{2}{*}{$.009$\tiny{$\pm.003$}}&\multirow{2}{*}{$.013$\tiny{$\pm.004$}}&\multirow{2}{*}{$.973$\tiny{$\pm.011$}}&\multirow{2}{*}{$1.616$\tiny{$\pm.102$}}\\
&&&&&&&\\
AdaGVAE&\multirow{2}{*}{$1.000$\tiny{$\pm.000$}}&\multirow{2}{*}{$.656$\tiny{$\pm.137$}}&\multirow{2}{*}{$.016$\tiny{$\pm.005$}}&\multirow{2}{*}{$.020$\tiny{$\pm.009$}}&\multirow{2}{*}{$.009$\tiny{$\pm.004$}}&\multirow{2}{*}{$.973$\tiny{$\pm.042$}}&\multirow{2}{*}{$1.620$\tiny{$\pm.147$}}\\
&&&&&&&\\

AdaMLVAE&\multirow{2}{*}{$.997$\tiny{$\pm.008$}}&\multirow{2}{*}{$.706$\tiny{$\pm.168$}}&\multirow{2}{*}{$.017$\tiny{$\pm.007$}}&\multirow{2}{*}{$.019$\tiny{$\pm.009$}}&\multirow{2}{*}{$.011$\tiny{$\pm.004$}}&\multirow{2}{*}{$.943$\tiny{$\pm.111$}}&\multirow{2}{*}{$1.395$\tiny{$\pm.117$}}\\
&&&&&&&\\

Quessard&\multirow{2}{*}{$1.000$\tiny{$\pm.000$}}&\multirow{2}{*}{$.596$\tiny{$\pm.032$}}&\multirow{2}{*}{$.016$\tiny{$\pm.006$}}&\multirow{2}{*}{$.008$\tiny{$\pm.004$}}&\multirow{2}{*}{$.017$\tiny{$\pm.008$}}&\multirow{2}{*}{$.999$\tiny{$\pm.000$}}&\multirow{2}{*}{$1.183$\tiny{$\pm.412$}}\\
&&&&&&&\\

\methodname{}&\multirow{2}{*}{$1.000$\tiny{$\pm.001$}}&\multirow{2}{*}{$.664$\tiny{$\pm.105$}}&\multirow{2}{*}{$.016$\tiny{$\pm.002$}}&\multirow{2}{*}{$.009$\tiny{$\pm.004$}}&\multirow{2}{*}{$.019$\tiny{$\pm.005$}}&\multirow{2}{*}{$.897$\tiny{$\pm.108$}}&\multirow{2}{*}{$1.627$\tiny{$\pm.104$}}\\
/0&&&&&&&\\
\methodname{}&\multirow{2}{*}{$1.000$\tiny{$\pm.000$}}&\multirow{2}{*}{$.662$\tiny{$\pm.046$}}&\multirow{2}{*}{$.017$\tiny{$\pm.005$}}&\multirow{2}{*}{$.009$\tiny{$\pm.004$}}&\multirow{2}{*}{$.020$\tiny{$\pm.005$}}&\multirow{2}{*}{$.963$\tiny{$\pm.010$}}&\multirow{2}{*}{$1.475$\tiny{$\pm.121$}}\\
/256&&&&&&&\\
\methodname{}&\multirow{2}{*}{$1.000$\tiny{$\pm.000$}}&\multirow{2}{*}{$.956$\tiny{$\pm.119$}}&\multirow{2}{*}{$.021$\tiny{$\pm.006$}}&\multirow{2}{*}{$.297$\tiny{$\pm.157$}}&\multirow{2}{*}{$.023$\tiny{$\pm.003$}}&\multirow{2}{*}{$.967$\tiny{$\pm.092$}}&\multirow{2}{*}{$.245$\tiny{$\pm.474$}}\\
/512&&&&&&&\\
\methodname{}&\multirow{2}{*}{$1.000$\tiny{$\pm.000$}}&\multirow{2}{*}{$1.000$\tiny{$\pm.000$}}&\multirow{2}{*}{$.022$\tiny{$\pm.006$}}&\multirow{2}{*}{$.390$\tiny{$\pm.022$}}&\multirow{2}{*}{$.026$\tiny{$\pm.003$}}&\multirow{2}{*}{$.999$\tiny{$\pm.000$}}&\multirow{2}{*}{$.000$\tiny{$\pm.000$}}\\
/768&&&&&&&\\
\methodname{}&\multirow{2}{*}{$1.000$\tiny{$\pm.000$}}&\multirow{2}{*}{$1.000$\tiny{$\pm.000$}}&\multirow{2}{*}{$.022$\tiny{$\pm.003$}}&\multirow{2}{*}{$.396$\tiny{$\pm.026$}}&\multirow{2}{*}{$.026$\tiny{$\pm.006$}}&\multirow{2}{*}{$.999$\tiny{$\pm.000$}}&\multirow{2}{*}{$.000$\tiny{$\pm.000$}}\\
/1024&&&&&&&\\
\methodname{}&\multirow{2}{*}{$1.000$\tiny{$\pm.000$}}&\multirow{2}{*}{$1.000$\tiny{$\pm.000$}}&\multirow{2}{*}{$.019$\tiny{$\pm.005$}}&\multirow{2}{*}{$.401$\tiny{$\pm.018$}}&\multirow{2}{*}{$.026$\tiny{$\pm.004$}}&\multirow{2}{*}{$.999$\tiny{$\pm.000$}}&\multirow{2}{*}{$.000$\tiny{$\pm.000$}}\\
/1280&&&&&&&\\
\methodname{}&\multirow{2}{*}{$1.000$\tiny{$\pm.000$}}&\multirow{2}{*}{$1.000$\tiny{$\pm.000$}}&\multirow{2}{*}{$.019$\tiny{$\pm.005$}}&\multirow{2}{*}{$.397$\tiny{$\pm.017$}}&\multirow{2}{*}{$.026$\tiny{$\pm.007$}}&\multirow{2}{*}{$.999$\tiny{$\pm.000$}}&\multirow{2}{*}{$.000$\tiny{$\pm.000$}}\\
/1536&&&&&&&\\
\methodname{}&\multirow{2}{*}{$1.000$\tiny{$\pm.000$}}&\multirow{2}{*}{$1.000$\tiny{$\pm.000$}}&\multirow{2}{*}{$.020$\tiny{$\pm.004$}}&\multirow{2}{*}{$.399$\tiny{$\pm.018$}}&\multirow{2}{*}{$.026$\tiny{$\pm.004$}}&\multirow{2}{*}{$.999$\tiny{$\pm.000$}}&\multirow{2}{*}{$.000$\tiny{$\pm.000$}}\\
/1792&&&&&&&\\
\methodname{}&\multirow{2}{*}{$1.000$\tiny{$\pm.000$}}&\multirow{2}{*}{$1.000$\tiny{$\pm.000$}}&\multirow{2}{*}{$.020$\tiny{$\pm.006$}}&\multirow{2}{*}{$.444$\tiny{$\pm.186$}}&\multirow{2}{*}{$.027$\tiny{$\pm.004$}}&\multirow{2}{*}{$.999$\tiny{$\pm.000$}}&\multirow{2}{*}{$.000$\tiny{$\pm.000$}}\\
/full&&&&&&&\\
\methodname{}&&&&&&&\multirow{2}{*}{$.016$\tiny{$\pm.006$}}\\
 /paths&&&&&&&\\
\bottomrule
\end{tabular}
\end{sc}
\end{small}
\end{center}
\vskip -0.1in
\end{table*}

\begin{table*}[!t]
\caption{Scores for the Airplane dataset.}
\label{tab:results_airplane}
\vskip 0.1in
\begin{center}
\begin{small}
\begin{sc}
\begin{tabular}{lccccccc}
\toprule
Model & Beta $\uparrow$ & Factor $\uparrow$ & SAP $\uparrow$ & DCI $\uparrow$ & MIG $\uparrow$ & MOD $\uparrow$ & \metricname{} $\downarrow$\\
\midrule
VAE&\multirow{2}{*}{$1.000$\tiny{$\pm.001$}}&\multirow{2}{*}{$.947$\tiny{$\pm.054$}}&\multirow{2}{*}{$.023$\tiny{$\pm.005$}}&\multirow{2}{*}{$.013$\tiny{$\pm.005$}}&\multirow{2}{*}{$.020$\tiny{$\pm.017$}}&\multirow{2}{*}{$.801$\tiny{$\pm.045$}}&\multirow{2}{*}{$1.342$\tiny{$\pm.084$}}\\
&&&&&&&\\
$\beta$-VAE&\multirow{2}{*}{$1.000$\tiny{$\pm.001$}}&\multirow{2}{*}{$.997$\tiny{$\pm.005$}}&\multirow{2}{*}{$.018$\tiny{$\pm.005$}}&\multirow{2}{*}{$.036$\tiny{$\pm.012$}}&\multirow{2}{*}{$.028$\tiny{$\pm.012$}}&\multirow{2}{*}{$.816$\tiny{$\pm.104$}}&\multirow{2}{*}{$1.481$\tiny{$\pm.129$}}\\
&&&&&&&\\
cc-VAE&\multirow{2}{*}{$.858$\tiny{$\pm.194$}}&\multirow{2}{*}{$.646$\tiny{$\pm.353$}}&\multirow{2}{*}{$.010$\tiny{$\pm.006$}}&\multirow{2}{*}{$.021$\tiny{$\pm.011$}}&\multirow{2}{*}{$.018$\tiny{$\pm.009$}}&\multirow{2}{*}{$.969$\tiny{$\pm.034$}}&\multirow{2}{*}{$1.481$\tiny{$\pm.174$}}\\
&&&&&&&\\
Factor&\multirow{2}{*}{$1.000$\tiny{$\pm.000$}}&\multirow{2}{*}{$.984$\tiny{$\pm.015$}}&\multirow{2}{*}{$.020$\tiny{$\pm.003$}}&\multirow{2}{*}{$.021$\tiny{$\pm.008$}}&\multirow{2}{*}{$.026$\tiny{$\pm.013$}}&\multirow{2}{*}{$.810$\tiny{$\pm.040$}}&\multirow{2}{*}{$1.382$\tiny{$\pm.171$}}\\
&&&&&&&\\
DIP-I&\multirow{2}{*}{$1.000$\tiny{$\pm.000$}}&\multirow{2}{*}{$.994$\tiny{$\pm.008$}}&\multirow{2}{*}{$.022$\tiny{$\pm.004$}}&\multirow{2}{*}{$.029$\tiny{$\pm.012$}}&\multirow{2}{*}{$.026$\tiny{$\pm.012$}}&\multirow{2}{*}{$.842$\tiny{$\pm.073$}}&\multirow{2}{*}{$1.289$\tiny{$\pm.150$}}\\
&&&&&&&\\
DIP-II&\multirow{2}{*}{$.998$\tiny{$\pm.005$}}&\multirow{2}{*}{$.972$\tiny{$\pm.031$}}&\multirow{2}{*}{$.021$\tiny{$\pm.004$}}&\multirow{2}{*}{$.022$\tiny{$\pm.013$}}&\multirow{2}{*}{$.030$\tiny{$\pm.019$}}&\multirow{2}{*}{$.780$\tiny{$\pm.054$}}&\multirow{2}{*}{$1.367$\tiny{$\pm.129$}}\\
&&&&&&&\\
AdaGVAE&\multirow{2}{*}{$.962$\tiny{$\pm.120$}}&\multirow{2}{*}{$.892$\tiny{$\pm.314$}}&\multirow{2}{*}{$.013$\tiny{$\pm.009$}}&\multirow{2}{*}{$.026$\tiny{$\pm.016$}}&\multirow{2}{*}{$.010$\tiny{$\pm.008$}}&\multirow{2}{*}{$.733$\tiny{$\pm.264$}}&\multirow{2}{*}{$1.029$\tiny{$\pm.288$}}\\
&&&&&&&\\
AdaMLVAE&\multirow{2}{*}{$1.000$\tiny{$\pm.000$}}&\multirow{2}{*}{$.995$\tiny{$\pm.007$}}&\multirow{2}{*}{$.019$\tiny{$\pm.009$}}&\multirow{2}{*}{$.035$\tiny{$\pm.011$}}&\multirow{2}{*}{$.017$\tiny{$\pm.009$}}&\multirow{2}{*}{$.861$\tiny{$\pm.073$}}&\multirow{2}{*}{$.994$\tiny{$\pm.275$}}\\
&&&&&&&\\
Quessard&\multirow{2}{*}{$.999$\tiny{$\pm.003$}}&\multirow{2}{*}{$.987$\tiny{$\pm.026$}}&\multirow{2}{*}{$.018$\tiny{$\pm.007$}}&\multirow{2}{*}{$.016$\tiny{$\pm.009$}}&\multirow{2}{*}{$.018$\tiny{$\pm.005$}}&\multirow{2}{*}{$.795$\tiny{$\pm.107$}}&\multirow{2}{*}{$.558$\tiny{$\pm.239$}}\\
&&&&&&&\\
\methodname{}&\multirow{2}{*}{$.536$\tiny{$\pm.065$}}&\multirow{2}{*}{$.000$\tiny{$\pm.000$}}&\multirow{2}{*}{$.002$\tiny{$\pm.001$}}&\multirow{2}{*}{$.007$\tiny{$\pm.004$}}&\multirow{2}{*}{$.005$\tiny{$\pm.003$}}&\multirow{2}{*}{$.956$\tiny{$\pm.046$}}&\multirow{2}{*}{$1.165$\tiny{$\pm.180$}}\\
/0&&&&&&&\\
\methodname{}&\multirow{2}{*}{$1.000$\tiny{$\pm.000$}}&\multirow{2}{*}{$1.000$\tiny{$\pm.000$}}&\multirow{2}{*}{$.022$\tiny{$\pm.006$}}&\multirow{2}{*}{$.144$\tiny{$\pm.011$}}&\multirow{2}{*}{$.023$\tiny{$\pm.004$}}&\multirow{2}{*}{$.870$\tiny{$\pm.039$}}&\multirow{2}{*}{$.153$\tiny{$\pm.021$}}\\
/256&&&&&&&\\
\methodname{}&\multirow{2}{*}{$1.000$\tiny{$\pm.000$}}&\multirow{2}{*}{$1.000$\tiny{$\pm.000$}}&\multirow{2}{*}{$.023$\tiny{$\pm.008$}}&\multirow{2}{*}{$.151$\tiny{$\pm.015$}}&\multirow{2}{*}{$.020$\tiny{$\pm.004$}}&\multirow{2}{*}{$.846$\tiny{$\pm.032$}}&\multirow{2}{*}{$.168$\tiny{$\pm.022$}}\\
/512&&&&&&&\\
\methodname{}&\multirow{2}{*}{$1.000$\tiny{$\pm.000$}}&\multirow{2}{*}{$1.000$\tiny{$\pm.000$}}&\multirow{2}{*}{$.022$\tiny{$\pm.004$}}&\multirow{2}{*}{$.140$\tiny{$\pm.014$}}&\multirow{2}{*}{$.022$\tiny{$\pm.005$}}&\multirow{2}{*}{$.832$\tiny{$\pm.034$}}&\multirow{2}{*}{$.180$\tiny{$\pm.030$}}\\
/768&&&&&&&\\
\methodname{}&\multirow{2}{*}{$1.000$\tiny{$\pm.000$}}&\multirow{2}{*}{$1.000$\tiny{$\pm.000$}}&\multirow{2}{*}{$.020$\tiny{$\pm.005$}}&\multirow{2}{*}{$.160$\tiny{$\pm.015$}}&\multirow{2}{*}{$.022$\tiny{$\pm.005$}}&\multirow{2}{*}{$.859$\tiny{$\pm.032$}}&\multirow{2}{*}{$.165$\tiny{$\pm.021$}}\\
/1024&&&&&&&\\
\methodname{}&\multirow{2}{*}{$1.000$\tiny{$\pm.000$}}&\multirow{2}{*}{$1.000$\tiny{$\pm.000$}}&\multirow{2}{*}{$.024$\tiny{$\pm.004$}}&\multirow{2}{*}{$.153$\tiny{$\pm.013$}}&\multirow{2}{*}{$.022$\tiny{$\pm.003$}}&\multirow{2}{*}{$.876$\tiny{$\pm.016$}}&\multirow{2}{*}{$.151$\tiny{$\pm.015$}}\\
/1280&&&&&&&\\
\methodname{}&\multirow{2}{*}{$1.000$\tiny{$\pm.000$}}&\multirow{2}{*}{$1.000$\tiny{$\pm.000$}}&\multirow{2}{*}{$.021$\tiny{$\pm.005$}}&\multirow{2}{*}{$.160$\tiny{$\pm.016$}}&\multirow{2}{*}{$.022$\tiny{$\pm.004$}}&\multirow{2}{*}{$.896$\tiny{$\pm.025$}}&\multirow{2}{*}{$.140$\tiny{$\pm.018$}}\\
/1536&&&&&&&\\
\methodname{}&\multirow{2}{*}{$1.000$\tiny{$\pm.000$}}&\multirow{2}{*}{$1.000$\tiny{$\pm.000$}}&\multirow{2}{*}{$.022$\tiny{$\pm.005$}}&\multirow{2}{*}{$.163$\tiny{$\pm.022$}}&\multirow{2}{*}{$.023$\tiny{$\pm.003$}}&\multirow{2}{*}{$.904$\tiny{$\pm.016$}}&\multirow{2}{*}{$.138$\tiny{$\pm.010$}}\\
/1792&&&&&&&\\
\methodname{}&\multirow{2}{*}{$1.000$\tiny{$\pm.000$}}&\multirow{2}{*}{$1.000$\tiny{$\pm.000$}}&\multirow{2}{*}{$.016$\tiny{$\pm.008$}}&\multirow{2}{*}{$.161$\tiny{$\pm.024$}}&\multirow{2}{*}{$.021$\tiny{$\pm.006$}}&\multirow{2}{*}{$.913$\tiny{$\pm.018$}}&\multirow{2}{*}{$.132$\tiny{$\pm.009$}}\\
/full&&&&&&&\\

\methodname{}&&&&&&&\multirow{2}{*}{$.185$\tiny{$\pm.017$}}\\
 /paths&&&&&&&\\
\bottomrule
\end{tabular}
\end{sc}
\end{small}
\end{center}
\vskip -0.1in
\end{table*}

\begin{table*}[!t]
\caption{Scores for the Modelnet40 Airplanes dataset.}
\label{tab:results_modelnet40}
\vskip 0.1in
\begin{center}
\begin{small}
\begin{sc}
\begin{tabular}{lccccccc}
\toprule
Model & Beta $\uparrow$ & Factor $\uparrow$ & SAP $\uparrow$ & DCI $\uparrow$ & MIG $\uparrow$ & MOD $\uparrow$ & \metricname{} $\downarrow$\\
\midrule
VAE&\multirow{2}{*}{$.995$\tiny{$\pm.004$}}&\multirow{2}{*}{$.838$\tiny{$\pm.030$}}&\multirow{2}{*}{$.013$\tiny{$\pm.002$}}&\multirow{2}{*}{$.013$\tiny{$\pm.002$}}&\multirow{2}{*}{$.009$\tiny{$\pm.002$}}&\multirow{2}{*}{$.415$\tiny{$\pm.058$}}&\multirow{2}{*}{$.393$\tiny{$\pm.110$}}\\
&&&&&&&\\
$\beta$-VAE&\multirow{2}{*}{$.995$\tiny{$\pm.005$}}&\multirow{2}{*}{$.857$\tiny{$\pm.045$}}&\multirow{2}{*}{$.012$\tiny{$\pm.003$}}&\multirow{2}{*}{$.015$\tiny{$\pm.003$}}&\multirow{2}{*}{$.009$\tiny{$\pm.002$}}&\multirow{2}{*}{$.447$\tiny{$\pm.067$}}&\multirow{2}{*}{$.285$\tiny{$\pm.045$}}\\
&&&&&&&\\
cc-VAE&\multirow{2}{*}{$.997$\tiny{$\pm.003$}}&\multirow{2}{*}{$.818$\tiny{$\pm.093$}}&\multirow{2}{*}{$.011$\tiny{$\pm.003$}}&\multirow{2}{*}{$.017$\tiny{$\pm.004$}}&\multirow{2}{*}{$.011$\tiny{$\pm.003$}}&\multirow{2}{*}{$.567$\tiny{$\pm.063$}}&\multirow{2}{*}{$.281$\tiny{$\pm.191$}}\\
&&&&&&&\\
Factor&\multirow{2}{*}{$.996$\tiny{$\pm.004$}}&\multirow{2}{*}{$.856$\tiny{$\pm.052$}}&\multirow{2}{*}{$.012$\tiny{$\pm.002$}}&\multirow{2}{*}{$.014$\tiny{$\pm.003$}}&\multirow{2}{*}{$.010$\tiny{$\pm.003$}}&\multirow{2}{*}{$.444$\tiny{$\pm.077$}}&\multirow{2}{*}{$.388$\tiny{$\pm.096$}}\\
&&&&&&&\\
DIP-I&\multirow{2}{*}{$.988$\tiny{$\pm.009$}}&\multirow{2}{*}{$.783$\tiny{$\pm.070$}}&\multirow{2}{*}{$.012$\tiny{$\pm.002$}}&\multirow{2}{*}{$.013$\tiny{$\pm.002$}}&\multirow{2}{*}{$.008$\tiny{$\pm.001$}}&\multirow{2}{*}{$.343$\tiny{$\pm.082$}}&\multirow{2}{*}{$.416$\tiny{$\pm.142$}}\\
&&&&&&&\\
DIP-II&\multirow{2}{*}{$.994$\tiny{$\pm.006$}}&\multirow{2}{*}{$.832$\tiny{$\pm.042$}}&\multirow{2}{*}{$.013$\tiny{$\pm.003$}}&\multirow{2}{*}{$.014$\tiny{$\pm.003$}}&\multirow{2}{*}{$.011$\tiny{$\pm.002$}}&\multirow{2}{*}{$.433$\tiny{$\pm.080$}}&\multirow{2}{*}{$.379$\tiny{$\pm.130$}}\\
&&&&&&&\\
AdaGVAE&\multirow{2}{*}{$.996$\tiny{$\pm.006$}}&\multirow{2}{*}{$.775$\tiny{$\pm.079$}}&\multirow{2}{*}{$.010$\tiny{$\pm.006$}}&\multirow{2}{*}{$.014$\tiny{$\pm.006$}}&\multirow{2}{*}{$.013$\tiny{$\pm.004$}}&\multirow{2}{*}{$.421$\tiny{$\pm.092$}}&\multirow{2}{*}{$.476$\tiny{$\pm.218$}}\\
&&&&&&&\\

AdaMLVAE&\multirow{2}{*}{$.996$\tiny{$\pm.006$}}&\multirow{2}{*}{$.784$\tiny{$\pm.055$}}&\multirow{2}{*}{$.012$\tiny{$\pm.006$}}&\multirow{2}{*}{$.014$\tiny{$\pm.005$}}&\multirow{2}{*}{$.014$\tiny{$\pm.004$}}&\multirow{2}{*}{$.445$\tiny{$\pm.040$}}&\multirow{2}{*}{$.580$\tiny{$\pm.141$}}\\
&&&&&&&\\
Quessard&\multirow{2}{*}{$.907$\tiny{$\pm.192$}}&\multirow{2}{*}{$.727$\tiny{$\pm.384$}}&\multirow{2}{*}{$.010$\tiny{$\pm.005$}}&\multirow{2}{*}{$.015$\tiny{$\pm.007$}}&\multirow{2}{*}{$.009$\tiny{$\pm.004$}}&\multirow{2}{*}{$.563$\tiny{$\pm.108$}}&\multirow{2}{*}{$.134$\tiny{$\pm.294$}}\\
&&&&&&&\\
\methodname{}&\multirow{2}{*}{$.990$\tiny{$\pm.009$}}&\multirow{2}{*}{$.863$\tiny{$\pm.038$}}&\multirow{2}{*}{$.011$\tiny{$\pm.003$}}&\multirow{2}{*}{$.015$\tiny{$\pm.003$}}&\multirow{2}{*}{$.014$\tiny{$\pm.003$}}&\multirow{2}{*}{$.538$\tiny{$\pm.103$}}&\multirow{2}{*}{$.731$\tiny{$\pm.068$}}\\
/0&&&&&&&\\
\methodname{}&\multirow{2}{*}{$1.000$\tiny{$\pm.000$}}&\multirow{2}{*}{$.990$\tiny{$\pm.004$}}&\multirow{2}{*}{$.012$\tiny{$\pm.005$}}&\multirow{2}{*}{$.052$\tiny{$\pm.009$}}&\multirow{2}{*}{$.020$\tiny{$\pm.006$}}&\multirow{2}{*}{$.947$\tiny{$\pm.007$}}&\multirow{2}{*}{$.041$\tiny{$\pm.007$}}\\
/full&&&&&&&\\
\bottomrule
\end{tabular}
\end{sc}
\end{small}
\end{center}
\vskip -0.1in
\end{table*}

\begin{table*}[!t]
\caption{Scores for COIL 100 dataset.}
\label{tab:results_coil100}
\vskip 0.1in
\begin{center}
\begin{small}
\begin{sc}
\begin{tabular}{lccccccc}
\toprule
Model & Beta $\uparrow$ & Factor $\uparrow$ & SAP $\uparrow$ & DCI $\uparrow$ & MIG $\uparrow$ & MOD $\uparrow$ & \metricname{} $\downarrow$\\
\midrule
VAE&\multirow{2}{*}{$1.000$\tiny{$\pm.000$}}&\multirow{2}{*}{$.674$\tiny{$\pm.049$}}&\multirow{2}{*}{$.014$\tiny{$\pm.003$}}&\multirow{2}{*}{$.016$\tiny{$\pm.003$}}&\multirow{2}{*}{$.011$\tiny{$\pm.002$}}&\multirow{2}{*}{$.986$\tiny{$\pm.001$}}&\multirow{2}{*}{$.463$\tiny{$\pm.030$}}\\
&&&&&&&\\
$\beta$-VAE&\multirow{2}{*}{$1.000$\tiny{$\pm.001$}}&\multirow{2}{*}{$.740$\tiny{$\pm.024$}}&\multirow{2}{*}{$.015$\tiny{$\pm.004$}}&\multirow{2}{*}{$.014$\tiny{$\pm.004$}}&\multirow{2}{*}{$.013$\tiny{$\pm.003$}}&\multirow{2}{*}{$.982$\tiny{$\pm.001$}}&\multirow{2}{*}{$.579$\tiny{$\pm.095$}}\\
&&&&&&&\\
cc-VAE&\multirow{2}{*}{$.999$\tiny{$\pm.003$}}&\multirow{2}{*}{$.723$\tiny{$\pm.026$}}&\multirow{2}{*}{$.013$\tiny{$\pm.005$}}&\multirow{2}{*}{$.014$\tiny{$\pm.003$}}&\multirow{2}{*}{$.013$\tiny{$\pm.004$}}&\multirow{2}{*}{$.985$\tiny{$\pm.001$}}&\multirow{2}{*}{$.406$\tiny{$\pm.057$}}\\
&&&&&&&\\
Factor&\multirow{2}{*}{$1.000$\tiny{$\pm.001$}}&\multirow{2}{*}{$.684$\tiny{$\pm.041$}}&\multirow{2}{*}{$.014$\tiny{$\pm.002$}}&\multirow{2}{*}{$.012$\tiny{$\pm.002$}}&\multirow{2}{*}{$.013$\tiny{$\pm.004$}}&\multirow{2}{*}{$.984$\tiny{$\pm.001$}}&\multirow{2}{*}{$.490$\tiny{$\pm.024$}}\\
&&&&&&&\\
DIP-I&\multirow{2}{*}{$.999$\tiny{$\pm.002$}}&\multirow{2}{*}{$.631$\tiny{$\pm.025$}}&\multirow{2}{*}{$.013$\tiny{$\pm.004$}}&\multirow{2}{*}{$.012$\tiny{$\pm.002$}}&\multirow{2}{*}{$.010$\tiny{$\pm.002$}}&\multirow{2}{*}{$.986$\tiny{$\pm.001$}}&\multirow{2}{*}{$.525$\tiny{$\pm.109$}}\\
&&&&&&&\\
DIP-II&\multirow{2}{*}{$1.000$\tiny{$\pm.001$}}&\multirow{2}{*}{$.643$\tiny{$\pm.043$}}&\multirow{2}{*}{$.013$\tiny{$\pm.003$}}&\multirow{2}{*}{$.014$\tiny{$\pm.002$}}&\multirow{2}{*}{$.011$\tiny{$\pm.002$}}&\multirow{2}{*}{$.985$\tiny{$\pm.001$}}&\multirow{2}{*}{$.568$\tiny{$\pm.079$}}\\
&&&&&&&\\
AdaGVAE&\multirow{2}{*}{$1.000$\tiny{$\pm.000$}}&\multirow{2}{*}{$.672$\tiny{$\pm.021$}}&\multirow{2}{*}{$.015$\tiny{$\pm.007$}}&\multirow{2}{*}{$.016$\tiny{$\pm.005$}}&\multirow{2}{*}{$.014$\tiny{$\pm.006$}}&\multirow{2}{*}{$.984$\tiny{$\pm.001$}}&\multirow{2}{*}{$.431$\tiny{$\pm.049$}}\\
&&&&&&&\\
AdaMLVAE&\multirow{2}{*}{$1.000$\tiny{$\pm.000$}}&\multirow{2}{*}{$.688$\tiny{$\pm.027$}}&\multirow{2}{*}{$.011$\tiny{$\pm.003$}}&\multirow{2}{*}{$.015$\tiny{$\pm.006$}}&\multirow{2}{*}{$.018$\tiny{$\pm.009$}}&\multirow{2}{*}{$.984$\tiny{$\pm.002$}}&\multirow{2}{*}{$.400$\tiny{$\pm.076$}}\\
&&&&&&&\\
Quessard&\multirow{2}{*}{$1.000$\tiny{$\pm.000$}}&\multirow{2}{*}{$.780$\tiny{$\pm.044$}}&\multirow{2}{*}{$.014$\tiny{$\pm.004$}}&\multirow{2}{*}{$.014$\tiny{$\pm.002$}}&\multirow{2}{*}{$.011$\tiny{$\pm.003$}}&\multirow{2}{*}{$.973$\tiny{$\pm.004$}}&\multirow{2}{*}{$.396$\tiny{$\pm.055$}}\\
&&&&&&&\\
\methodname{}&\multirow{2}{*}{$1.000$\tiny{$\pm.001$}}&\multirow{2}{*}{$.739$\tiny{$\pm.047$}}&\multirow{2}{*}{$.014$\tiny{$\pm.003$}}&\multirow{2}{*}{$.014$\tiny{$\pm.001$}}&\multirow{2}{*}{$.011$\tiny{$\pm.001$}}&\multirow{2}{*}{$.982$\tiny{$\pm.004$}}&\multirow{2}{*}{$.515$\tiny{$\pm.099$}}\\
/0&&&&&&&\\
\methodname{}&\multirow{2}{*}{$1.000$\tiny{$\pm.000$}}&\multirow{2}{*}{$.655$\tiny{$\pm.028$}}&\multirow{2}{*}{$.015$\tiny{$\pm.004$}}&\multirow{2}{*}{$.029$\tiny{$\pm.003$}}&\multirow{2}{*}{$.013$\tiny{$\pm.003$}}&\multirow{2}{*}{$.802$\tiny{$\pm.056$}}&\multirow{2}{*}{$.112$\tiny{$\pm.026$}}\\
/full&&&&&&&\\
\bottomrule
\end{tabular}
\end{sc}
\end{small}
\end{center}
\vskip -0.1in
\end{table*}


\end{document}